\documentclass[11pt]{article} 
\usepackage{url}
\usepackage{smile}
\usepackage{graphicx} 
\usepackage{epstopdf}
\usepackage{wrapfig}
\usepackage[colorlinks, linkcolor=blue, anchorcolor=blue, citecolor=blue]{hyperref}

\usepackage[margin=1in]{geometry}
\usepackage[normalem]{ulem}
\usepackage[export]{adjustbox}
\usepackage{mathtools, cuted}
\usepackage{natbib}
\usepackage{enumerate}
\usepackage{enumitem}
\usepackage[utf8]{inputenc} 
\usepackage[T1]{fontenc}    
\usepackage{hyperref}       
\usepackage{url}            
\usepackage{booktabs}       
\usepackage{amsfonts}       
\usepackage{nicefrac}       
\usepackage{microtype}      
\usepackage{xcolor}         
\usepackage{graphicx}
\usepackage{hyperref}

\usepackage{amsmath}
\usepackage{amssymb}
\usepackage{mathtools}
\usepackage{amsthm}
\usepackage{multirow}
\usepackage{colortbl}
\usepackage{pifont}
\usepackage{subcaption}
\usepackage{algorithm, algorithmic}
\usepackage{wrapfig}

\usepackage{kpfonts}
\DeclareMathAlphabet{\mathsf}{OT1}{cmss}{m}{n}

\SetMathAlphabet{\mathsf}{bold}{OT1}{cmss}{bx}{n}

\newcommand{\R}{\mathbb{R}}
\newcommand{\nn}{n}
\newcommand{\nh}{H}
\newcommand{\nng}{n_{g}}
\newcommand{\nbu}{n_{b}}
\newcommand{\dd}{d}
\newcommand{\ddh}{d_{H}}
\newcommand{\bb}{b}
\newcommand{\rs}{s}
\newcommand{\rr}{r}
\newcommand{\gs}{g}
\newcommand{\bx}{\bm{x}}
\newcommand{\bq}{\bm{q}}
\newcommand{\bk}{\bm{k}}
\newcommand{\bv}{\bm{v}}
\newcommand{\bxt}{\bx_{t}}
\newcommand{\bqt}{\bq_{t}}
\newcommand{\bkt}{\bk_{t}}
\newcommand{\bvt}{\bv_{t}}
\newcommand{\bmu}{\bm{u}}
\newcommand{\bmm}{\bm{m}}
\newcommand{\mX}{\bm{X}}
\newcommand{\mH}{\bm{H}}
\newcommand{\mW}{\bm{W}}
\newcommand{\mQ}{\bm{Q}}
\newcommand{\mK}{\bm{K}}
\newcommand{\mV}{\bm{V}}
\newcommand{\mD}{\bm{D}}
\newcommand{\mL}{\bm{L}}
\newcommand{\mLh}{\bm{L}_{h}}
\newcommand{\mS}{\bm{S}}
\newcommand{\mA}{\bm{A}}
\newcommand{\mAh}{\bm{A}_{h}}
\newcommand{\mB}{\bm{B}}
\newcommand{\mBh}{\bm{B}_{h}}
\newcommand{\mR}{\bm{R}}
\newcommand{\mRh}{\bm{R}_{h}}

\newcommand{\Wo}{\mW_{o}}
\newcommand{\Wqi}{\mW_{q_{i}}}
\newcommand{\Wki}{\mW_{k_{i}}}
\newcommand{\Wvi}{\mW_{v_{i}}}

\newcommand{\mQi}{\mQ^{(i)}}
\newcommand{\mKi}{\mK^{(i)}}
\newcommand{\mVi}{\mV^{(i)}}

\newcommand{\mKt}{\mK_{t}}
\newcommand{\mVt}{\mV_{t}}
\newcommand{\mXq}{\widehat{\mX}}
\newcommand{\mDq}{\widehat{\mD}}

\newcommand{\Gcali}{\mathcal{G}_{i}}
\newcommand{\Bcal}{\mathcal{B}}

\newcommand{\Quant}{\textrm{Quant}}
\newcommand{\Filter}{\textrm{Filter}}
\newcommand{\SVDSolver}{\textrm{SVDSolver}}

\newcommand{\ouralg}{GEAR}

\author{Hao Kang$^{*}$, Qingru Zhang$^{*}$, Souvik Kundu, Geonhwa Jeong,\\ Zaoxing Liu, Tushar Krishna, Tuo Zhao$^\dagger$
}

\title{GEAR: An Efficient KV Cache Compression Recipe \\ for Near-Lossless Generative Inference of LLM}

\begin{document}

\maketitle

\def\thefootnote{$\dagger$}\footnotetext{Hao Kang, Qingru Zhang, Geonhwa Jeong, Tushar Krishna, and Tuo Zhao are affiliated with Georgia Tech. Souvik Kundu is affiliated with Intel. Zaoxing Liu is affiliated with the University of Maryland. Correspondence to \url{hkang342@gatech.edu}, \url{qingru.zhang@gatech.edu}, \url{souvikk.kundu@intel.com}, and \url{tourzhao@gatech.edu}.}
\def\thefootnote{*}\footnotetext{Equal contributions}

\begin{abstract}

Key-value (KV) caching has become the de-facto technique to accelerate generation speed for large language models (LLMs) inference. However, the growing cache demand with increasing sequence length has transformed LLM inference to be a \textit{memory bound} problem, significantly constraining the system throughput. Existing methods rely on dropping unimportant tokens or quantizing entries group-wise. Such methods, however, often incur high approximation errors to represent the compressed matrices. The autoregressive decoding process further compounds the error of each step, resulting in critical deviation in model generation and deterioration of performance. To tackle this challenge, we propose {\ouralg}, an efficient \textit{error reduction framework} that augments a quantization scheme with two error reduction components and achieves near-lossless performance at high compression ratios. {\ouralg} first applies quantization to majority of entries of similar magnitudes to ultra-low precision. It then employs a low-rank matrix to approximate the quantization error, and a sparse matrix to remedy individual errors from outlier entries.  By adeptly integrating three techniques, {\ouralg} is able to fully exploit their synergistic potentials. Our experiments show that {\ouralg} can maintain similar accuracy to that of FP16 cache with improvement up to $24.42\%$ over the SOTA baselines at 2-bit compression. Additionally, compared to LLM inference with FP16 KV cache, {\ouralg} can reduce peak-memory of up to $2.39\times$, bringing $2.1\times\sim 5.07\times$ throughput improvement. Our code is publicly available at. \url{https://github.com/HaoKang-Timmy/GEAR}.

\end{abstract}

\section{Introduction}\label{sec:introduction}
Autoregressive large language models (LLMs) \citep{GPT3,zhang2022opt,touvron2023llama,touvron2023llama2} have marked a significant milestone in natural language processing (NLP) and artificial intelligence (AI) \citep{vaswani2017attention,brown2020language,openai2023gpt4}, showcasing exceptional performances across a wide range of applications, such as content creation and dialogue systems \citep{yuan2022wordcraft,thoppilan2022lamda,wei2022emergent}. 
When serving these LLMs for generative inference, {\it KV cache}-ing has become a routine practice. It stores previously computed Key/Value tensors from attention calculation and reuses them while generating next tokens \citep{pope2022efficiently}, avoiding intensive recalculation to improve the generation speed.  

Despite its prominence, the memory consumption of the KV cache grows rapidly with the model size and sequence length, imposing significant constraints on system throughput. 
For instance, in the case of a 30 billion-parameter LLM with an input length of 1024 and batch size of 128, the resulting KV cache can occupy up to 180 GB of memory \citep{zhang2023h2o}. 
To alleviate this pressure on limited GPU memory capacity,  inference systems typically resort to {\it offloading} \citep{aminabadi2022deepspeed,sheng2023flexgen} -- transferring the KV cache to CPU memory or NVMe storage. This, however, can still introduce non-trivial overhead due to the limited PCIe bandwidth between GPUs and CPUs on many devices. 
Therefore, it is crucial to reduce the intensive memory footprint of the emerging bottleneck of KV cache in generative inference.

\begin{figure*}[t!]
	\centering
	\begin{subfigure}{0.32\textwidth}
		\centering
		\includegraphics[width=0.85\textwidth]{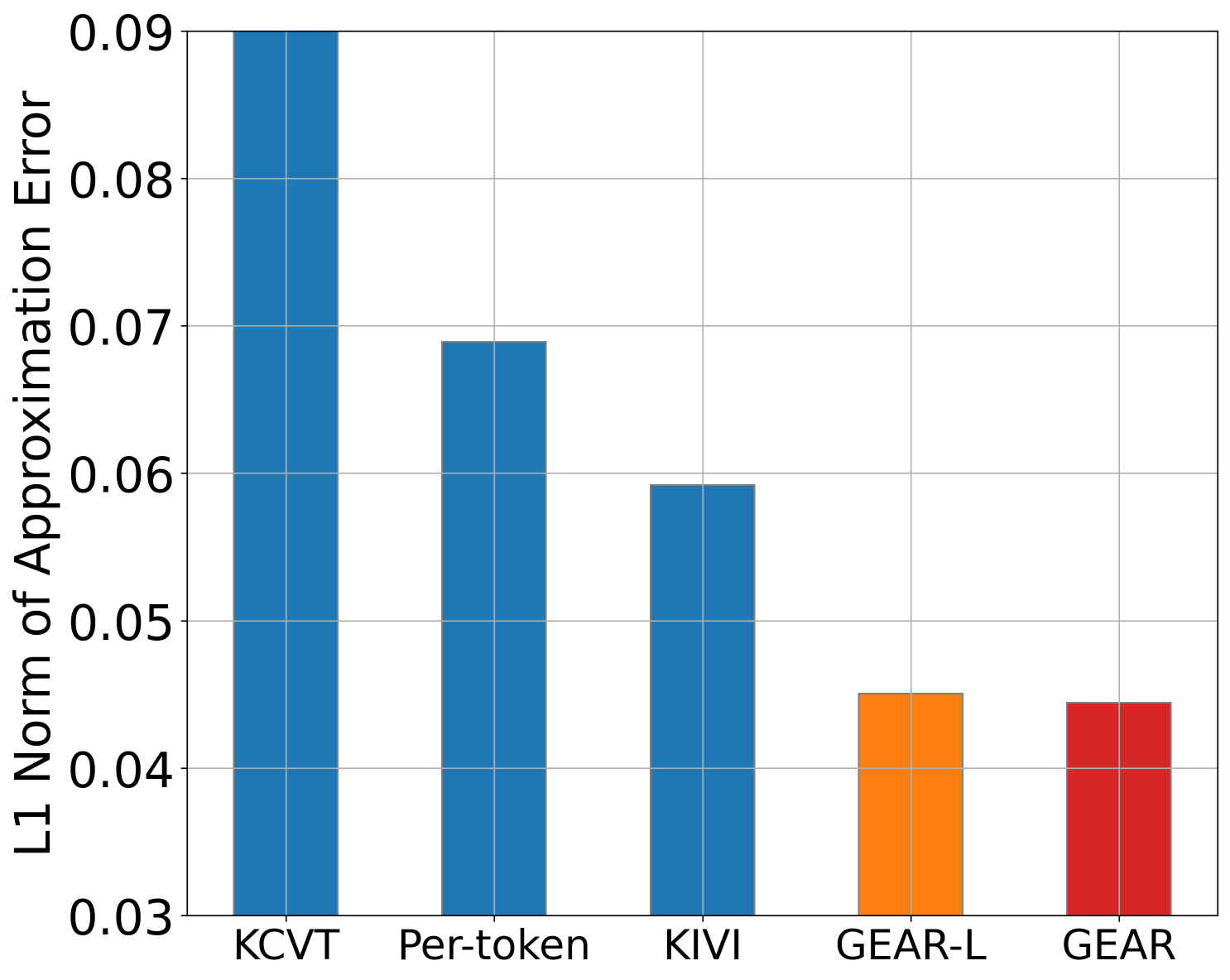}
		\caption{\small Approx. error on GSM8k-CoT}
		\label{fig:approximation_error_gsm8k}
	\end{subfigure}
    \begin{subfigure}{0.32\textwidth}
		\centering
		\includegraphics[width=0.862\textwidth]{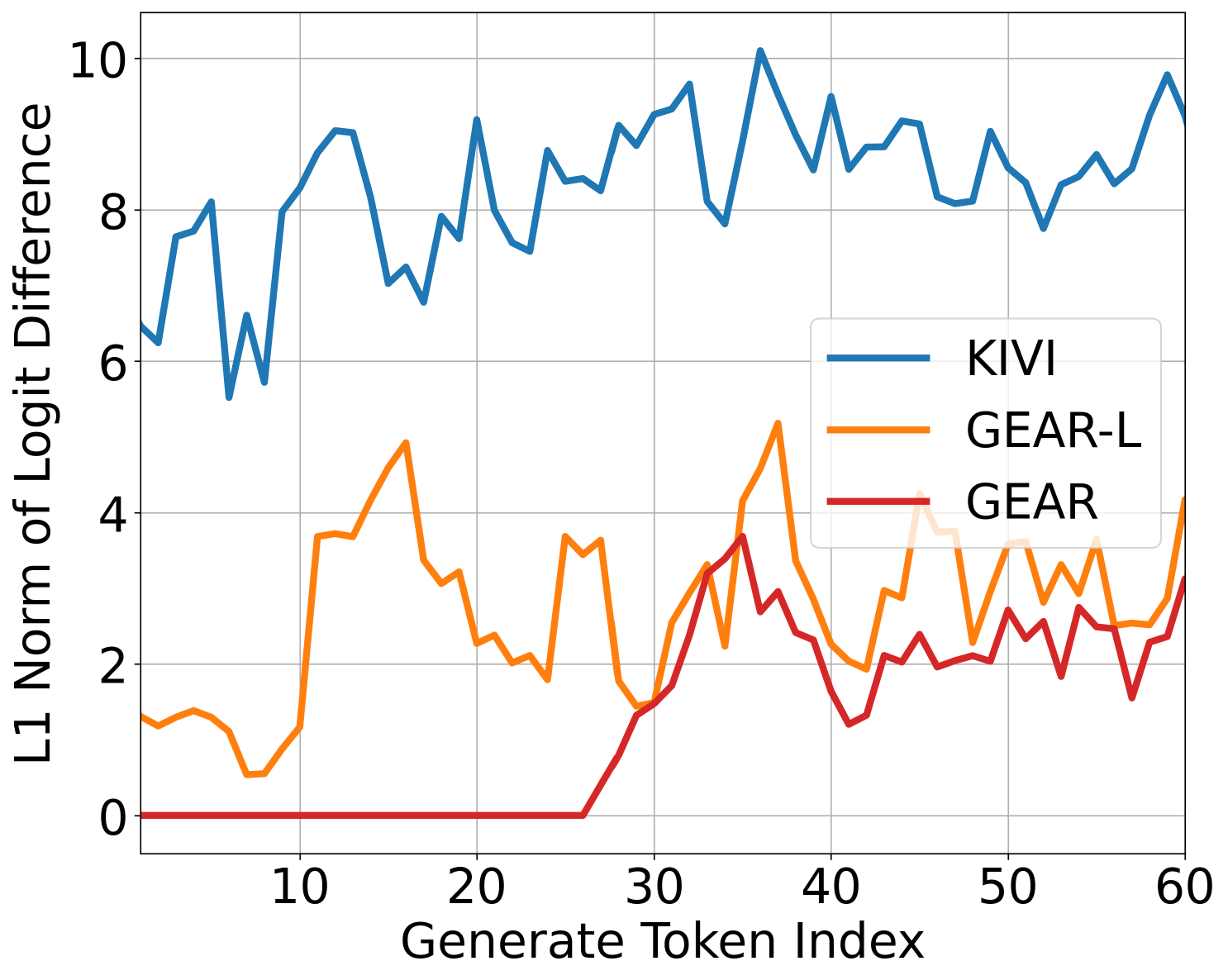}
		\caption{\small Difference in prediction logits}
		\label{fig:generation_deviation}
	\end{subfigure}
	\begin{subfigure}{0.32\textwidth}
		\centering
		\includegraphics[width=0.88\textwidth]{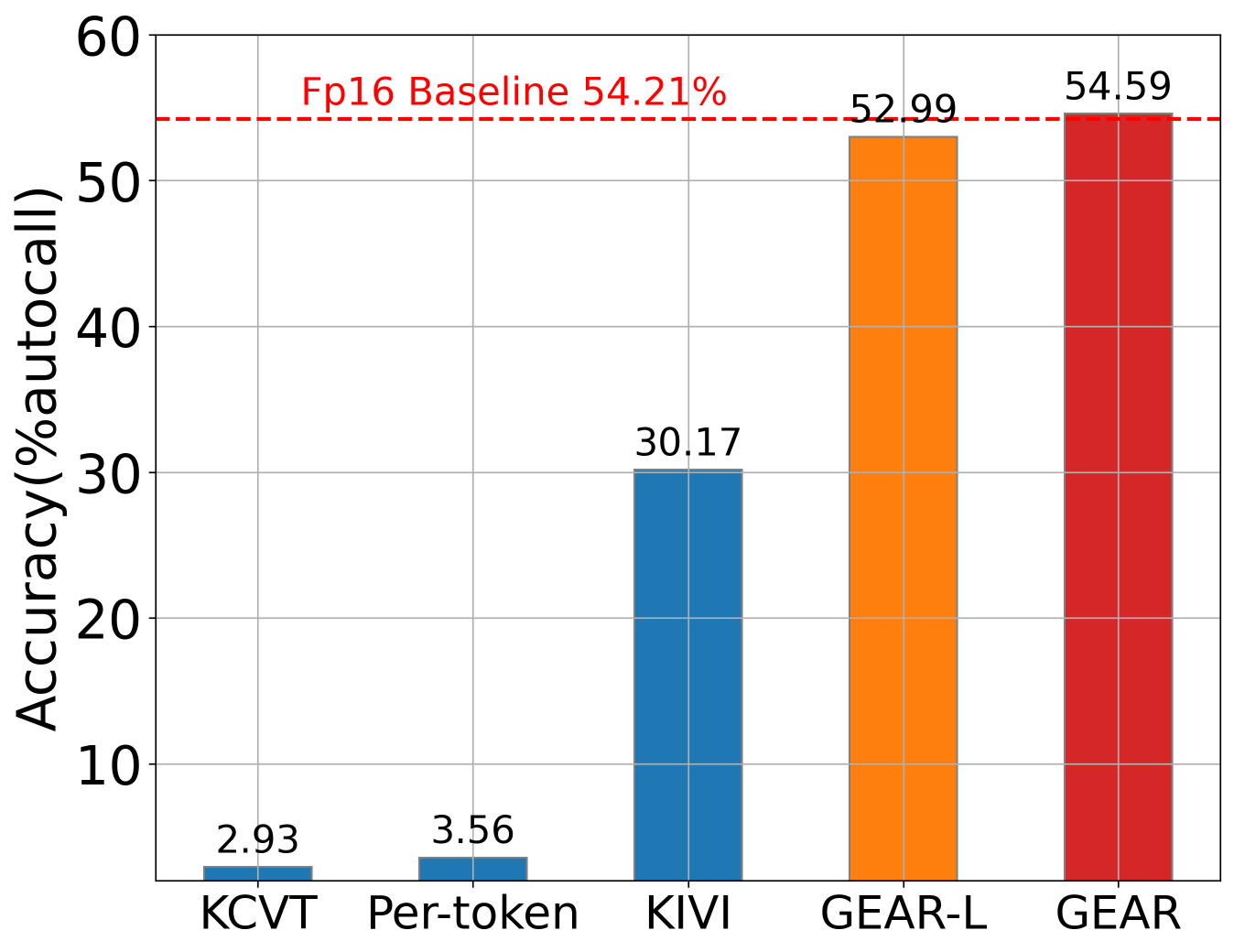}
		\caption{\small Accuracy on GSM8k-CoT}
		\label{fig:acc_gsm8k_cot}
	\end{subfigure}
	\caption{\small (\ref{fig:approximation_error_gsm8k}) compares the approximation error when compressing KV caches to 2-bit for LLaMA3-8B on GSM8k (w.~CoT). (\ref{fig:generation_deviation}) presents difference in prediction logits from FP16 baseline after compressing KV caches of an GSM8k (w.~CoT) example, indicating the approximation error can be severely compounded along steps and critically divert model generations. (\ref{fig:acc_gsm8k_cot}) shows reducing the error can significantly improve the performance. 
    }
	\label{fig:introduction}
\end{figure*}

To address this issue, {\it token dropping} methods have been proposed to compress the cache size while maintaining the generative performance \citep{zhang2023h2o,liu2023scissorhands,ge2023model}. 
These approaches harness the sparsity observed in attention scores to evict embeddings of less important tokens from the KV cache while retaining frequently attended ones. 
For example, H\textsubscript{2}O \citep{zhang2023h2o} utilizes accumulated attention scores to evaluate token importance and reduces cache size by dropping tokens with lower scores.  
In addition, {\it quantization} is another widely-adopted compression scheme that maps full-precision tensor values into discrete levels and store them at lower precision, e.g.,~INT4 or INT8 \citep{Zafrir_2019,dettmers2022llm,sheng2023flexgen}.  
For example, FlexGen \citep{sheng2023flexgen} employs a fine-grained group-wise asymmetric quantization that groups KV entries per-token, divides $\gs$ contiguous entries as a group, and quantize the tensor group-wise. 
Two concurrent works \citep{liu2024kivi,hooper2024kvquant} further study KV entry distribution and propose to quantize Key cache per-channel and quantize Value cache per-token, compressing the cache size by a high ratio. 

The existing methods can effectively compress the cache size to low-precision while achieving near-lossless performance on natural language understanding tasks like multiple-choice QA, text classification, or simple summarization task \citep{zhang2023h2o,liu2024kivi}. 
However, a stark contrast emerges when applying these methods to complex generative tasks that require models to generate longer responses or involve reasoning, such as mathematical problem-solving \citep{cobbe2021training} and chain-of-thought (CoT) reasoning \citep{wei2023chainofthought}. Their performance deteriorates under a high compression ratio\footnote{We define the compression ratio as tensor size in FP16 divided by that in compressed format.} (e.g.,~4-bit/2-bit quantization or dropping $>50\%$ tokens \citep{ge2023model}), which is noticeable in both types of methods\footnote{Please refer to Section~\ref{sec:experiments} for our empirical evidence.}. 
This phenomenon can be attributed to the non-trivial approximation error induced by them, i.e.,~difference between original KV and the compressed ones. 
For simple tasks, models are required to generate only few tokens where necessary information for correct prediction can often be derived from a small set of important contextual tokens. 
Consequently, a relatively large approximation error does not significantly hinder the generation of target tokens. 
In contrast, the complex tasks require models to generate longer sequences conditioned on prompts that often contains densely correlated information (e.g.,~CoT reasoning). The autoregressive decoding can compound the approximation error at every step. Consequently, the negative effect of even a relatively small error can be magnified along generation steps, adversely affecting subsequent generation.  
As an example, Figure~\ref{fig:introduction} presents the approximation error of various methods on GSM8k and illustrates the deviation in token generations due to the accumulated error, which degrades the accuracy a lot. 
Therefore, the crux of the issue lies in high approximation errors of these methods, especially under high compression ratios.




To address this challenge, we propose {\it{\ouralg}} (\underline{GE}nerative Inference with \underline{A}pproximation Error \underline{R}eduction), an efficient error reduction framework that augments existing KV cache quantization schemes with two error-reduction techniques, and adeptly integrate them to exploit their full potentials. 
Generally speaking, our framework consists of three components to decompose KV matrices: 
(i) First, we apply an existing {\it quantization method} to efficiently quantize the majority (e.g.,~98\%) of entries of similar magnitudes to ultra-low precision. 
(ii) Then, we introduce a {\it low-rank matrix} to efficiently approximate the quantization residuals. 
(iii) Finally, we employ a {\it sparse matrix} consisting of a negligible ratio of entries of large magnitudes to remedy the individual errors caused by these outliers. 
Such a composite approximation decouples the coherent parts from incoherent parts of the approximation error: the low-rank matrix captures the majority of coherent basis of quantization error while the sparse matrix rectifies the incoherency existing in individual outliers. Meanwhile, as shown by our empirical evidence in Section~\ref{sec:inference_efficiency}, these two lightweight components result in negligible memory and computational overheads, demonstrating high efficiency. As such, {\ouralg} can effectively reduce the approximation error in a highly efficient way and achieve superior performance on {\it both} complex and relatively simple tasks at high compression ratios in a {\it plug-and-play} manner. 
We find that using both sparse and low-rank components is necessary for {\ouralg} to establish the best performance, highlighting their complementary nature. 
Remarkably, for those prioritizing efficiency, equipping low-rank approximation alone for quantization can still effectively reduce the approximation error, yielding both significant efficiency and performance improvement. We refer to this lite version of GEAR as {\it GEAR-L}. 

Additionally, we incorporate a {\it streaming buffer} strategy for {\ouralg} to further improve inference efficiency. Specifically, when generating long sequences, we store KV vectors of newly generated tokens to a small buffer (e.g.,~buffer size $ \nbu = 20 $). When the buffer reaches its capacity, {\ouralg} conducts the KV cache compression every $ \nbu $ steps.  
As such, the inference speed can be significantly improved at a trivial cost of additional memory. 
Furthermore, to minimize the overhead, we demonstrate a GPU-friendly kernel implementation for {\ouralg}, which leverage the streaming and quantization benefit to improve inference throughputs significantly. 

We conduct experiments on diverse tasks and models to demonstrate the effectiveness of {\ouralg}. Specifically, we evaluate compression performance with LLaMA2-7B/13B \citep{touvron2023llama2}, Mistral-7B \citep{jiang2023mistral}, and LLaMA3-8B\citep{meta2024llama3} on generative tasks including mathematical reasoning (GSM8k\cite{cobbe2021training} and AQuA\citep{ling2017program}), symbolic reasoning (BigBench Hard\cite{suzgun2022challenging}), and long-context understanding (LongBench\cite{bai2023longbench}). 
We show that {\ouralg} consistently outperforms the baseline methods especially at high compression ratios such as 2-bit precision. 
For example, when compressing KV caches to 2-bit, {\ouralg} achieves an remarkable average accuracy improvement of {\bf 14.95}\% over the best-performing baseline across various models and datasets. 
Regarding the inference efficiency, compared to the FP16 baseline, {\ouralg} can reduce the peak memory up to {\bf2.39$\times$}, bring {\bf 2.10}$\times$ $\sim$ {\bf5.07}$\times$ throughput improvement. 
\section{Background}\label{sec:background}


{\bf Multi-head attention}. A typical transformer model consists of $ L $ stacked layers, where each layer contains two submodules: a multi-head attention (MHA) and a feed-forward network (FFN). 
Given the input token embeddings as $ \mX \in \R^{\nn\times \dd} $, MHA performs attention function in parallel $ \nh $ heads: 
\begin{equation}\label{eq:mha}
\text{MHA}\left(\mX \right) = \textrm{Concat}(\mH^{(1)},..., \mH^{(\nh)})\Wo, \quad \mH^{(i)} = \textrm{Softmax}\left( \mQi \mK^{(i)\top}/{\sqrt{\ddh}} \right) \mVi
\end{equation}
where $ \mQi = \mX\Wqi, \mKi = \mX\Wki, \mVi = \mX\Wvi $ are Query/Key/Value matrices, and $ \Wqi, \Wki, \Wvi \in \R^{\dd\times \ddh} $ are projection matrices of head $ i $. $ \ddh $ is typically set to $ \dd/\nh $. 

{\bf Prefill and decoding}. Suppose the model generates $\nng$ tokens.  
At the first generation step, the input tokens $\mX_{0}\in\R^{\nn\times \dd}$ are prefilled. Then $\mKi$ and $\mVi$ at every head and every layer are cached for subsequent generation, resulting in initial KV caches of prefill phrase: $\mK_{0} = \textrm{Concat}(\mK^{(1)}, \dots, \mK^{(\nh)})$, $\mV_{0} = \textrm{Concat}(\mV^{(1)}, \dots, \mV^{(\nh)})$ and $\mK_{0},\mV_{0} \in \R^{n\times \dd}$. At each step $t$ ($1\leq t\leq \nng$) of autoregressive decoding, the model predicts a new token $\bxt$ conditioned on the input and previously generated tokens.  At the following step, MHA only needs to compute the Query/Key/Value vectors\footnote{For simplicity, we concatenate multi-head embeddings here.} ($\bqt, \bkt, \bvt\in\R^{\dd}$) for the newly generated token $\bxt$ and appends $\bkt, \bvt$ to the KV cache: $\mK_{t} = \mK_{t-1} \| \bkt $, $\mV_{t} = \mV_{t-1}\|\bvt$. Then it performs the attention (\ref{eq:mha}) between $\bqt$ and $\mK_{t}$, $\mV_{t}$ only. 

{\bf Group-wise Quantization}. 
Group-wise quantization is widely applied to compress KV cache\cite{sheng2023flexgen,liu2024kivi,hooper2024kvquant}.
Given a tensor $\mX\in\R^{\nn\times\dd}$ in full precision, the vanilla method groups entries per-token by placing $\gs$ consecutive entries of a token into one group, e.g.,~the $i$-th group $\mX_{\Gcali}$ contains entries with indices $\Gcali = \{(t_i,c_i),\dots,(t_i,c_i+\gs) \}$ where $(t_i, c_i)$ is the beginning index of group $i$ and $\gs$ is group size. Then, it quantizes $\mX$ group-wise: $\mXq = \textrm{Quant}_{\bb,\gs}^{(\textrm{per-token})} $ with
\begin{align}\label{eq:group_quant}
& \textrm{Quant}_{\bb,\gs}^{(\textrm{per-token})}(\mX)_{\Gcali} = \left\lceil {(\mX_{\Gcali} - \min\mX_{\Gcali})}/{\Delta_{i}} \right\rfloor, \ \Delta_{i} = {(\max{\mX_{\Gcali}} - \min\mX_{\Gcali})}/{(2^{\bb}-1)}
\end{align}
where $\bb$ is the bit-width, $\mXq$ is the quantized tensor in $\bb$-bit precision, and $\lceil\cdot\rfloor$ is the rounding function. $\Delta_{i}$ and $\min\mX_{\Gcali}$ are the scaling factor and zero-point of group $i$. 
Two concurrent works (KIVI \cite{liu2024kivi} and KVQuant \cite{hooper2024kvquant}) explore the entry distribution of KV cache and show that, in Key cache, some fixed channels exhibit very large magnitudes. To confine the quantization error to each individual channel without impacting others, they propose to quantize Key cache per-channel while quantizing Value cache per-token, achieving the start-of-the-art 2-bit compression. 

Intuitively, more fine-grained grouping with smaller group size, such as $\gs=64$ in KIVI \cite{liu2024kivi}, leads to more accurate approximation and yields better performance. 
However, small group size induces considerable memory overheads due to the increased number of scaling factors and zero-points stored in full precision for each group. 
Meanwhile, fine-grained grouping for per-channel quantization leads to maintaining a residual subset of KV tokens in full precision until they form a complete group \cite{liu2024kivi}. Hence, the residual length of this full-precision parts should be set as a multiple of group size (e.g.,~128 as set by KIVI), further resulting in additional considerable overheads. 
To leverage the SOTA quantization scheme while minimizing overheads, we choose per-channel Key and per-token Value quantization without fine-grained grouping as a lite quantization backbone. We refer to it as {\it KCVT}, a variant of KIVI with coarse-grained per-vector grouping where all Key entries of one channel forms a group of size $\nn$ and all Value entries of one token forms a group of size $\dd$, significantly reducing the scaling and zero point storage overhead.

\section{GEAR Framework}\label{sec:method}

\begin{figure*}[t!]
	\vspace{-6mm}
	\centering
	\begin{subfigure}{0.32\textwidth}
		\centering
		\includegraphics[width=0.95\textwidth]{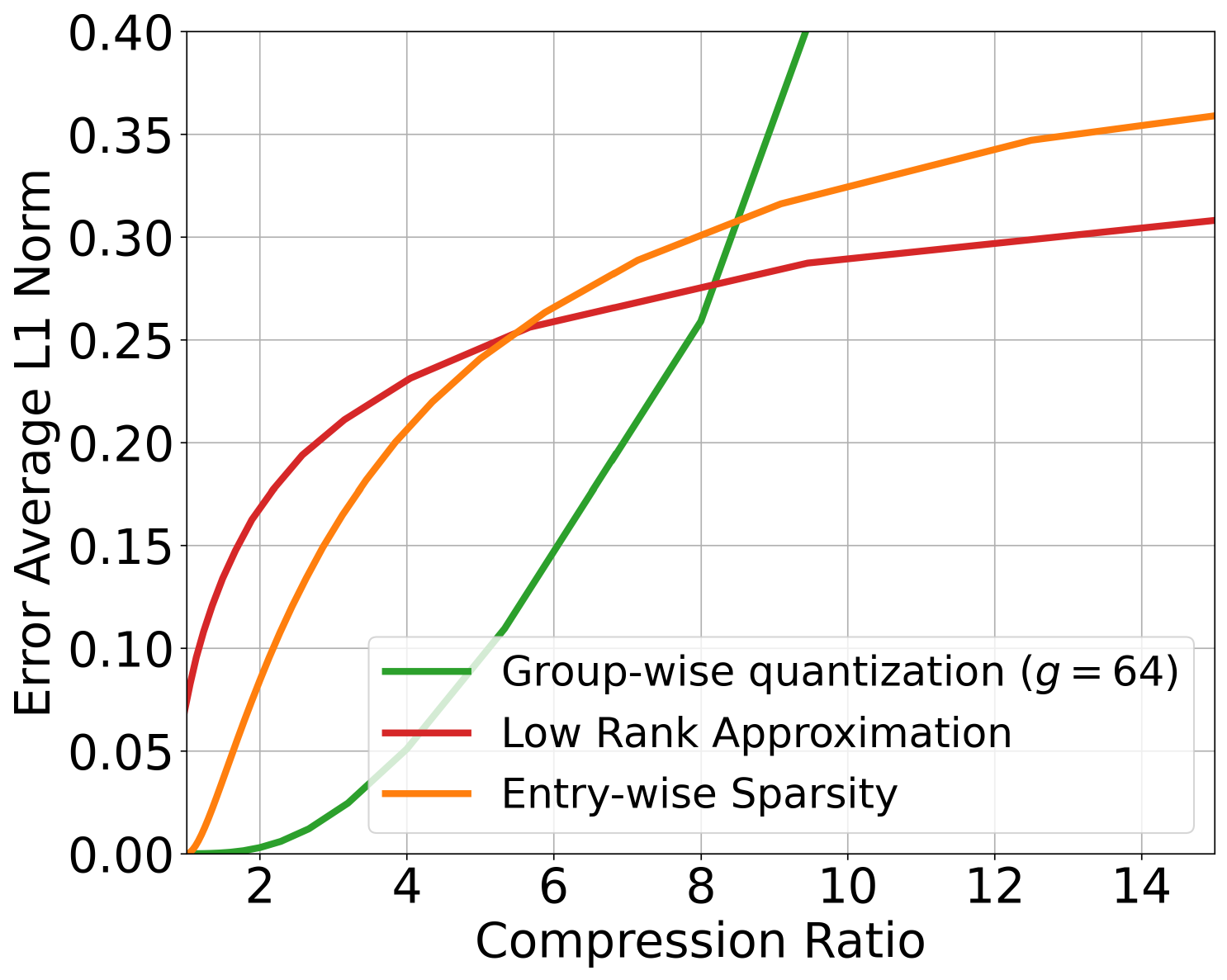}
		\caption{\small Error of each method}
		\label{fig:individual_error}
	\end{subfigure}
    \begin{subfigure}{0.32\textwidth}
		\centering
		\includegraphics[width=0.95\textwidth]{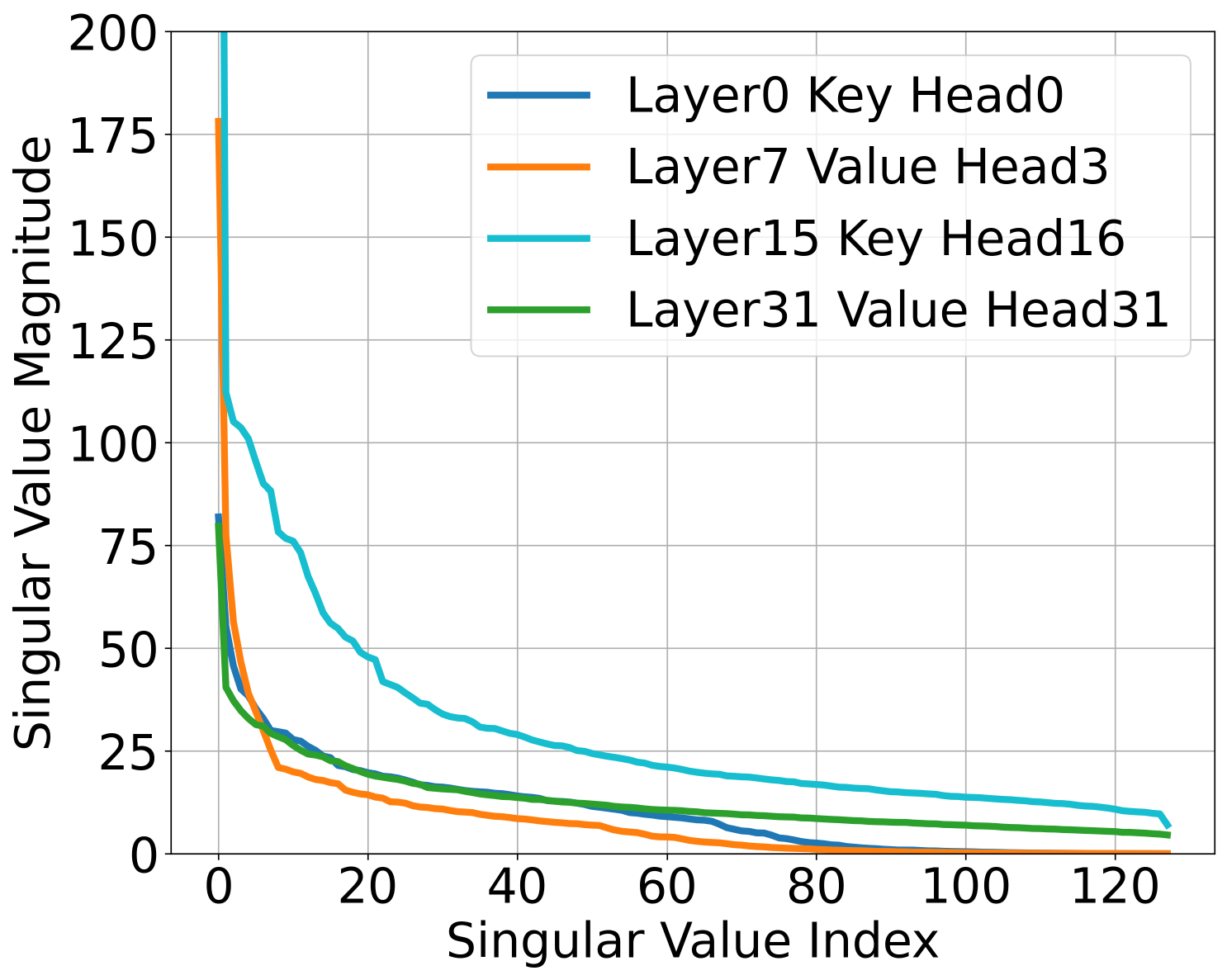}
		\caption{\small Spectrum of the residual}
		\label{fig:singular_value}
	\end{subfigure}
    \begin{subfigure}{0.32\textwidth}
		\centering
		\includegraphics[width=0.95\textwidth]{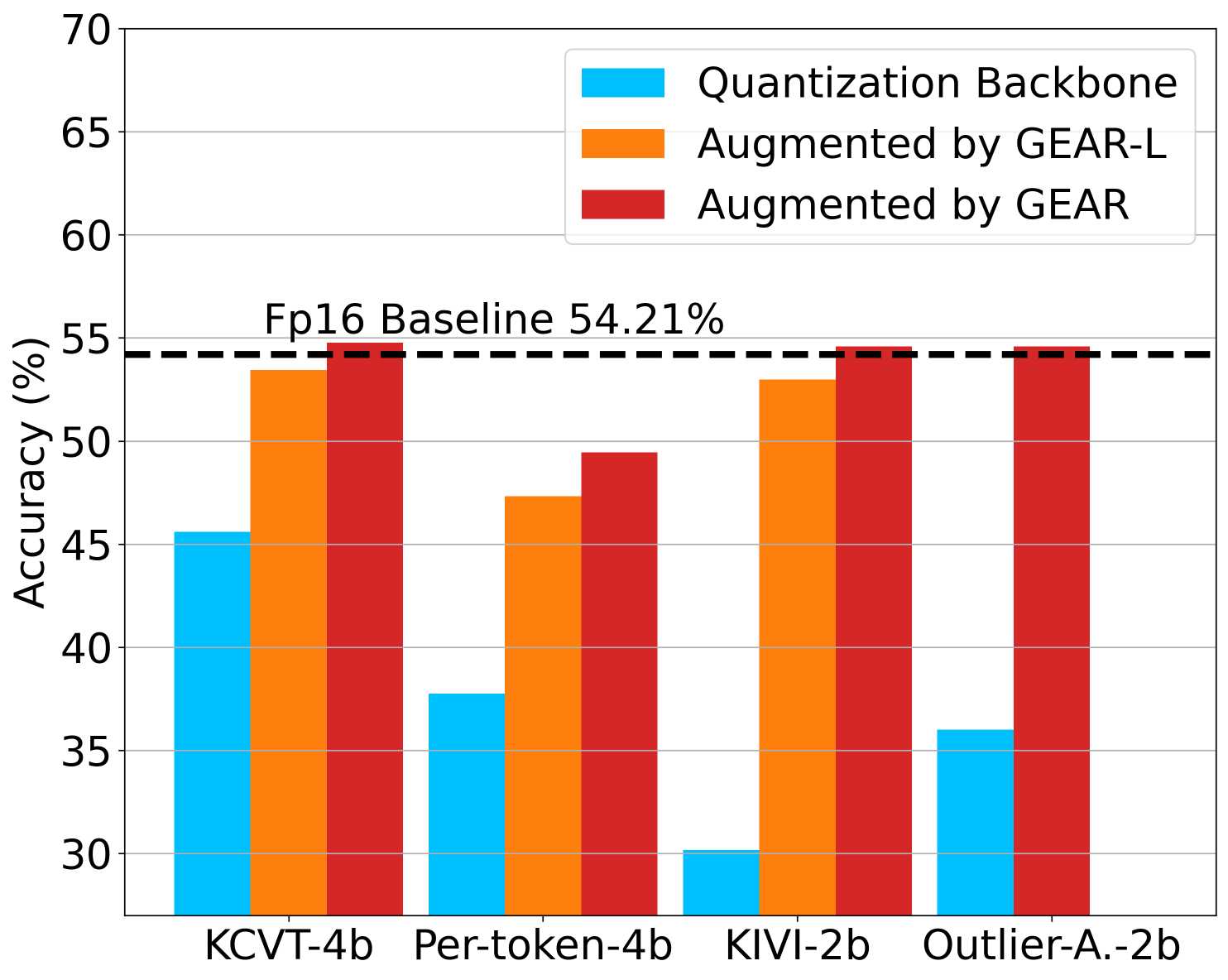}
		\caption{\small LLaMA3-8B on GSM8k-CoT}
		\label{fig:gear_augment_quant}
	\end{subfigure}
	\caption{\small (\ref{fig:individual_error}, \ref{fig:singular_value}) We randomly sample a GSM8k example and analyze its KV caches by LLaMA2-7B. (\ref{fig:individual_error}): the minimal approximation error of each individual technique when approximating the Value cache of the first layer; (\ref{fig:singular_value}): spectrum of the residual $\mRh$ decays rapidly. (\ref{fig:gear_augment_quant}): As an efficient error-reduction framework, GEAR is orthogonal to any off-the-shelf quantization and can augment them to achieve near-lossless accuracy. 
    }
	\label{fig:approximation_error}
\end{figure*}

The GEAR framework consists of three important components to decompose and compress a KV cache matrix: (i) a quantized matrix $ \mDq $ to serve as a compressed backbone; (ii) a low-rank matrix $ \mL $ to approximate the quantization residual; (iii) a sparse matrix $ \mS $ to capture the individual outliers. 

As discussed in Section~\ref{sec:introduction}, the approximation error plays a pivotal role in determining model performance. 
Therefore, given a tensor $ \mX\in\{\mKt,\mVt\} $, our objective is to minimize the error of approximating $ \mX $ with its compressed counterpart. A simple strategy is to employ each of three compression methods individually and approximate $ \mX $ by minimizing the distance to it. For instance, constructing $ \mL $ using the top singular values/vectors of $ \mX $ or composing $ \mS $ with the entries of largest magnitudes. However, as demonstrated by Figure~\ref{fig:individual_error}, solely relying on any of these three methods cannot achieve high compression ratios because they all result in substantially increased error under high compression ratios. 
Additionally, $ \mDq,\mL, \mS $ can function differently in the matrix approximation, capturing different components of $ \mX $. 
These motivations encourage us to explore the integration of three techniques to leverage their individual advantages while exploiting their synergistic potential. To achieve this, our goal becomes minimizing the following approximation error: 
\begin{align}\label{eq:approximation_error_target}
	\min_{\mDq, \mL, \mS} \left\lVert{\mX - \mDq - \mL - \mS} \right\rVert_{\rm F}. 
\end{align}
One interesting idea to minimize (\ref{eq:approximation_error_target}) is alternating among quantization, singular-value decomposition (SVD) and outlier extraction, and iteratively updating three matrices $ \mDq, \mL, \mS $ until achieving minimal error. This idea has been introduced by \citet{li2023loftq} to optimize a similar objective for an accurate initialization of weight quantization. However, the inference system has demanding speed requirements. The significant latency caused by these iterative updates is unacceptable for generative inference. Therefore, we propose an efficient solution to minimize the approximation error (\ref{eq:approximation_error_target}).

{\bf Outlier-aware quantization}. Inspired by the recent study on weight quantization \citep{kim2023squeezellm}, we observe that the quantized backbone $ \mDq $ and the sparse matrix $ \mS $ complement each other in the KV cache compression. Specifically, the quantization scheme can result in non-trivial quantization errors within each group due to the existence of outlier entries. 
Therefore, a straightforward strategy is to filter out these outlier before quantization. 
To align with grouping of per-channel Key and per-token Value quantization, we leverage a per-vector outlier filtering. 
Given an input tensor $\mX=\mKt$ (or $\mVt$), we extract both $\frac{\rs}{2} \%$ of maximum and minimum entries of each channel (or token) vector and store them in full precision with a sparse matrix $ \mS=\Filter_{\rs}(\mX) $ where
\begin{align}\label{eq:sparse_function}
	\begin{aligned}
		\Filter_{\rs}(\mX)_{ij} = \left\{
		\begin{array}{lc}
			\mX_{ij} &\textrm{if $\mX=\mKt$ and}~\mX_{ij}~\textrm{in top/bottom $\frac{\rs}{2}$\% of the $j$-th channel $\mX_{\cdot j}$}, \\
            \mX_{ij} &\textrm{if $\mX=\mVt$ and}~\mX_{ij}~\textrm{in top/bottom $\frac{\rs}{2}$\% of the $i$-th token $\mX_{i\cdot}$}, \\
			0 &\textrm{otherwise}.
		\end{array}
		\right.
	\end{aligned}
\end{align}
Then, we perform the quantization on the extracted matrix and obtain the quantized backbone: 
\begin{align}\label{eq:dense_quant}
	\mDq = \Quant_{\bb}^{(\textrm{Selected scheme})}(\mX - \mS). 
\end{align}
The outlier extraction technique has been applied by \citet{kim2023squeezellm} to augment training-dependent non-uniform weight quantization. 
In contrast to their application scenario, we explore the potential of outlier extraction techniques in conjunction with tuning-free uniform quantization for KV caches. 
It is important to note that, compared to weights, KV cache compression presents unique challenges because KV caches can contain more outliers, making its accurate quantization more challenging than weights \citep{xiao2023smoothquant}. 
Our empirical results in Section~\ref{sec:analysis_ablation} also show that the outlier-aware quantization faces challenges in achieving ultra-low precision compression such as 2-bit on complex generative tasks. 
To achieve effective high-ratio compression, it is often necessary to extract a large portion of outliers stored in a sparse matrix. 
However, representing such a sparse matrix with two index vectors and one value vector in full precision results in considerable memory overheads.
These suggest that, while using $\mS$ can reduce the error, it still falls short of fully remediating the error in an efficient way.




{\bf Low-rank approximation}. To reduce the approximation error more efficiently, we resort to {\it low-rank approximation}. 
As inspired by fact that various attention heads encode diverse contextual information within different channel ranges \cite{tenney2019bert,lift-prune,zhang2024tell}, we propose to apply {\it head-wise low-rank decomposition} on the residual $ \mR = \mX-(\mDq+\mS)  \in\R^{\nn\times\dd} $. Specifically, we first reshape $\mR$ along channel dimension and obtain $\nh$ multi-head sub-matrices $\{\mRh = \mR[:, (h-1)\ddh:h\ddh]\in\R^{\nn\times\ddh} | 1\leq h\leq \nh \}$ where $\mRh$ is the residual of head $h$. 
Suppose $\mRh$ has singular value decomposition as $ \sum_{i=1}^{k} \sigma_i \bmu_{i} \bmm_{i}^{\top}  $, where $ \sigma_1\geq\dots\geq\sigma_k $ are singular values and $ \bmu_i, \bmm_i $ are the corresponding singular vectors. 
As shown in Figure~\ref{fig:singular_value}, we empirically observe that the spectrum of residual matrices drops rapidly at the beginning. 
This suggests the existence of a coherent component within the residual. 
This component is represented by the top singular values/vectors, and shared among tokens, indicating the vector similarity. 
By these top singular values and vectors, we can efficiently capture and recover this coherent information, leading to an effective approximation to the quantization residual. 
To this end, we introduce a matrix $\mL = \textrm{Concat}(\mL_{1},\dots,\mL_{\nh})$, where $ \mLh $ is a low-rank matrix:  
\begin{align}\label{eq:low_rank_approximation}
	\mLh = \mAh\mBh^{\top} = \SVDSolver_{\rr}(\mRh)
\end{align}
$ \mAh\in\R^{\nn\times\rr}, \mBh\in\R^{\ddh\times\rr} $ and $ \rr $ is much smaller than $ \nn, \ddh $. For example, when $ \nn=1024 $ and $ \ddh=128 $, $ \rr=4 $ is sufficient to achieve near-lossless high-ratio compression. For $ \SVDSolver(\cdot) $, we employ an efficient {\it power iteration} algorithm \cite{powersgd}. This algorithm calculates $ \mAh, \mBh $ rapidly while ensuring that $ \mAh\mBh^{\top} $ accurately approximates the top-$ \rr $ singular values/vectors $ \sum_{i=1}^{r}\sigma_i\bmu_{i}\bmm_{i}^{\top} $ (please see Appendix~\ref{app:PI_algorithm} for the algorithm details). In the multi-batch setting, we apply low-rank approximation to input tensors batch-wise and head-wise.

In summary, {\ouralg} integrates three compression techniques to provide an efficient solution for minimizing the approximation error in (\ref{eq:approximation_error_target}). Specifically, the quantized backbone $ \mDq $ leverages the entry-wise similarity and compresses the majority of entries to the ultra-low precision. The low-rank matrix $ \mLh $ capitalizes on vector-wise similarity to extract the commonly shared information within the residuals. The sparse matrix $ \mS $ compensates for the extraction of sparse information existing in individual outliers and compliments the quantization process. 
As such, {\ouralg} effectively reduces the approximation error, achieving high-ratio KV cache compression. 
We recommend to use {\ouralg} with all three components for the best performance -- both near-lossless 4-bit and 2-bit performance as an alternate to SOTA methods. However, to prioritize efficiency, one can resort to a lite version of {\ouralg}, namely {\it GEAR-L}, that equips only low-rank approximation to restore quantization error, costing less memory-overhead while improving accuracy significantly.  
Finally, we highlight that, as an efficient error-reduction framework, GEAR(-L) is {\it orthogonal} to any off-the-shelf quantization scheme and can augment them to achieve near-lossless accuracy as shown in Figure~\ref{fig:gear_augment_quant} and Section~\ref{sec:experiments}. 

{\bf Streaming Buffer.} {\ouralg} also introduces a {{\it streaming buffer}} strategy during decoding to significantly boost its inference speed. Specifically, when serving the long-sequence generation, {\ouralg} stores KV vectors of newly generated tokens in full precision to a small buffer $ \Bcal $ of size $ \nbu $ (e.g.,~$\nbu=20$). When the buffer reaches its capacity every $ \nbu $ decoding steps, {\ouralg} conduct the compression for new tokens in $ \Bcal $ while the subsequent low-rank approximation is only performed on the new tokens. 
The concurrent work, KIVI \citep{liu2024kivi}, introduces a similar buffering approach to cache residual tokens until they complete a group. Hence, their residual buffer size should be set as a multiple of group size. 
In the case of coarse-grained grouping of KCVT, the buffer size can be set arbitrarily and we select a small size like $\nbu=20$ to enhance the inference speed while avoiding the non-trivial memory overheads. 
We summarize the detailed algorithm of {\ouralg} in Algorithm~{\ref{alg:our_algorithm}} of Appendix~\ref{app:gear_algorithm}.

\section{Experiments}\label{sec:experiments}

{
\setlength{\tabcolsep}{0.20em}
\renewcommand{\arraystretch}{1.0}
\begin{table*}[t!]
\vspace{-10mm}
\caption{Results on CoT reasoning tasks, which are hard generative task. Here, {\it KV Size} is the average $\%$ of the remaining size of compressed KV caches with respect to the size in FP16. The best results are shown in {\bf bold}. {\it N.A.} represents the extermely degenerated performance.}
\label{tab:main_cot_result}
\vspace{-3mm}
\begin{center}
\begin{small}
\begin{tabular}{l|c|c|ccc|ccc|ccc|c}
\toprule
\multicolumn{3}{c}{\bf Model} 
& \multicolumn{3}{|c}{\bf LLaMA3-8B} 
& \multicolumn{3}{|c|}{\bf LLaMA2-13B}
& \multicolumn{3}{|c}{\bf Mistral-7B}
& \multicolumn{1}{|c}{\bf All}
\\
\midrule
\multirow{2}*{\bf Method} 
& \multirow{1}*{\bf Bit} 
& \multirow{1}*{\bf Ave. }
& {\bf\footnotesize GSM8k} 
& {\bf\footnotesize AQuA} 
& {\bf\footnotesize BBH}
& {\bf\footnotesize GSM8k} 
& {\bf\footnotesize AQuA} 
& {\bf\footnotesize BBH}
& {\bf\footnotesize GSM8k} 
& {\bf\footnotesize AQuA} 
& {\bf\footnotesize BBH}
& {\bf\footnotesize Ave.}
\\
& {\bf $\bb$}
& { KV size}
& {Acc} 
& {Acc}
& {Acc}
& {Acc} 
& {Acc}
& {Acc}
& {Acc} 
& {Acc}
& {Acc}
& {Acc}
\\
\midrule
{FP16} 
& {16}
& {100\%}
& {54.21}
& {38.19}
& {53.66}
& {30.34}
& {21.65}
& {40.79}
& {42.84}
& {35.04}
& {47.92}
& {40.52}
\\
\midrule
\midrule
{\footnotesize Per-token Q.\textsubscript{$\gs=64$}}
& {4} 
& {34.2\%}
& {37.07}
& {39.37}
& {46.42}
& {20.85}
& {18.90}
& {34.72}
& {31.47}
& {29.13}
& {28.88}
& {31.94}
\\
{\footnotesize KCVT Quant}
& {4}
& {27.1\%}
& {45.59}
& {36.61}
& {51.67}
& {21.14}
& {21.05}
& {36.71}
& {30.31}
& {24.37}
& {46.86}
& {34.92}
\\
{\footnotesize KIVI\textsubscript{$\gs=64,\nbu=64$}}
& {4}
& {34.2\%}
& {46.25}
& {36.22}
& {48.03}
& {22.14}
& {21.65}
& {37.76}
& {32.83}
& {25.98}
& {44.56}
& {35.05}
\\
\midrule
{\footnotesize ${\textrm{\bf GEAR-L}}_{\rr=4}^{(\textrm{KCVT})}$ }
& {4}
& {29.0\%}
& {53.44} 
& {38.98} 
& {52.23} 
& {\textbf{30.25}} 
& {23.23} 
& {38.52}
& {\textbf{43.06}}
& {33.07}
& {47.42}
& {40.02}
\\
{\footnotesize ${\textrm{\bf\ouralg}}_{\rs=2\%,\rr=4}^{(\textrm{KCVT})}$ }
& {4}
& {31.0\%}
& {\bf54.76}
& {\textbf{40.55}} 
& {\textbf{52.74}} 
& {30.17}
& {\textbf{24.05}}
& {\textbf{40.63}}
& {41.93}
& {\textbf{34.57}}
& {\textbf{47.84}}
& {\bf40.80}
\\
\midrule
\midrule
{\footnotesize Per-token Q.\textsubscript{$\gs=64$}}
& {2} 
& {21.7\%}
& {3.56}
& {9.84}
& {4.72}
& {\it N.A.}
& {10.54}
& {\it N.A.}
& {\it N.A.}
& {11.42}
& {5.93}
& {7.67}
\\
{\footnotesize KIVI\textsubscript{$\gs=64,\nbu=64$}}
& {2}
& {21.7\%}
& {30.17}
& {25.36}
& {30.92}
& {16.60}
& {17.72} 
& {29.43}
& {23.35}
& {22.44}
& {31.28}
& {25.25}
\\
\midrule 
{\footnotesize ${\textrm{\bf GEAR-L}}_{\rr=4}^{(\textrm{KIVI})}$}
& {2}
& {23.6\%}
& {52.62} 
& {38.19} 
& {\textbf{51.44}} 
& {26.61} 
& {20.87} 
& {39.44} 
& {39.27}
& {29.92}
& {46.36}
& {38.34}
\\
{\footnotesize ${\textrm{\bf\ouralg}}_{\rs=2\%,\rr=4}^{(\textrm{KIVI},{\gs=64})}$}
& {2}
& {27.6\%}
& {\textbf{54.59}}
& {\textbf{38.19}}
& {50.30}
& {\textbf{30.27}}
& {\textbf{23.62}}
& {\textbf{39.67}}
& {\textbf{43.14}}
& {\textbf{33.96}}
& {\textbf{48.03}}
& {\bf40.20}
\\
\bottomrule
\end{tabular}
\end{small}
\end{center}
\vspace{-9mm}
\end{table*}
}

We use {\ouralg} as a {\it plug-and-play} KV cache compression for generative inference with various LLM models (including LLaMA2-7B/13B \cite{touvron2023llama2}, Mistral-7B \cite{jiang2023mistral} and LLaMA3-8B \cite{meta2024llama3}) on generative tasks including math reasoning (GSM8k \citep{cobbe2021training} and AQuA \citep{ling2017program}), symbolic reasoning (BigBench Hard (BBH) \citep{suzgun2022challenging}) with CoT prompting \cite{wei2023chainofthought}, and long-context understanding (LongBench \citep{bai2023longbench}).

{\bf Implementation optimization and details.}  
To minimize the overheads, we demonstrate via GPU kernel support and optimize the implementation for {\ouralg} as follows. Firstly, we fuse the dequantization with matrix multiplication using CUDA to improve decoding latency. Secondly, we integrate the streaming buffer for both the Key and Value such that newly generated Key/Value caches are all compressed every $\nbu$ steps. Moreover, due to streaming buffer during decoding, low-rank approximation is performed every $\nbu$ steps for only buffered tokens with ultra low rank ($\rr=2$), improving compression efficiency. Thirdly, we preform the forward pass of low-rank matrices on a separate path where down projection (e.g.,~$\bq_{h}^{\top}\mBh$) is first computed, followed by up projection (e.g.,~$(\bq_{h}^{\top}\mBh) \mAh^{\top}$), reducing computational complexity of their forward pass.  

We apply GEAR and baseline methods to open-source pre-trained LLMs available at \textit{Huggingface} \cite{wolf2019huggingface}, using our inference framework written in \textit{PyTorch} \cite{paszke2019pytorch}. 
As we focus on evaluating the impact of KV Cache compression,  we keep all other tensors in FP16, unless otherwise stated. 
We focus on ultra-low precision quantization and report the results of 4-bit and 2-bit quantization. 
For GEAR, we fix the sparsity ratio $\rs$ at 2\%, set the rank $\rr$ to 4 for inputs in prefill phase, and set the rank to 2 for each group of $\nbu$ new tokens in decoding phase. 
We find that the efficient KCVT quantization achieves effective 4-bit compression and hence leverage it as 4-bit quantization backbone for {\ouralg} due to its efficiency. However, in case of 2-bit compression, its performance degenerates a lot and the quantization schemes have to resort to fine-grained grouping to establish acceptable accuracy. Hence, we use KIVI as 2-bit quantization backbone for {\ouralg}. As demonstrated by \cite{liu2024kivi} that KIVI is not sensitive to group size $\gs$ and residual length $\nbu$ (Table 5 in \citep{liu2024kivi}), we thus select the group size as 64 and the residual length as 64 for both GEAR and KIVI in order to lower KV size overheads.
The superscript in bracket shown in Table \ref{tab:main_cot_result} and \ref{tab:main_result2} identifies the backbone quantization scheme. 


{\bf Baselines}. We compare {\ouralg} with the following baseline methods: 

$\bullet$ {\it Per-token group-wise quantization} (used in FlexGen \citep{sheng2023flexgen}) is a widely-adopted  method that quantizes KV cache per-token with fine-grained grouping. 

$\bullet$ {\it KIVI} \citep{liu2024kivi} is a concurrent KV cache quantization method that achieves start-of-the-art 2-bit KV cache compression. This method quantizes Key cache per-channel and quantizes Value cache per-token with fine-grained grouping, and stored residual tokens of length $\nbu$ in full precision. 

$\bullet$ {\it KCVT quantization} is a variant of KIVI that quantize Key cache per-channel and Value cache per-token without fine-grained grouping. This is a per-vector quantization that induces lower overheads. 

$\bullet$ {\it H\textsubscript{2}O} \citep{zhang2023h2o} is a recent token dropping method evicting unimportant tokens with lower accumulated attention scores, which we compare with in Table~\ref{tab:gsm8k_h20_gear}. 

Recent concurrent work KVQuant~\citep{hooper2024kvquant} explored KCVT to incorporate it with data-dependent non-uniform quantization. While it demonstrates near-lossless performance mainly on perplexity with WikiText2 and C4, it requires additional calibration that minimizes a Hessian-related objective driven by data samples to obtain the quantization signposts. 
GEAR, however, aims to be a plug-and-play method, that can be deployed with any inference quantization scheme without any such dependency. Thus we keep any calibration-dependent compression beyond the scope of this work. 

{
\setlength{\tabcolsep}{0.25em}
\renewcommand{\arraystretch}{1.1}
\begin{table*}[tb!]
\vspace{-8mm}
\caption{Results on GSM8k 5-shot and LongBench evaluation. Here, {\it KV Size} is the average $\%$ of the remaining size of compressed KV caches with respect to that in FP16 (i.e.,~the inverse of compression ratio). The best results are shown in {\bf bold}. Results marked as $\dagger$ are taken from other papers.  
}
\label{tab:main_result2}
\vspace{-2mm}
\begin{center}
\begin{small}
\begin{tabular}{l|c|ccc|ccc|cc}
\toprule
\multicolumn{2}{c|}{\bf Dataset}
& \multicolumn{3}{c|}{\bf GSM8k 5-shot}
& \multicolumn{5}{c}{\bf LongBench w. LLaMA2-7B}
\\
\midrule
\multirow{2}*{\bf Method} 
& \multirow{1}*{\bf Bit} 
& {\bf Ave. KV}
& {\bf 7B}
& {\bf 8B}
& {\bf\footnotesize QMSum}
& {\bf\footnotesize SAMSum}
& {\bf\footnotesize GovReport}
& \multicolumn{2}{c}{\bf 21 Tasks Ave.}
\\
& {\bf $\bb$}
& {\bf size}
& {Acc}
& {Acc}
& {Rouge}
& {Rouge}
& {Rouge}
& {Ave. KV }
& {Ave. score}
\\
\midrule
{FP16} 
& {16}
& {100\%}
& {13.50}
& {49.89}
& {21.28}
& {43.57}
& {26.06}
& {100\%}
& {26.82}
\\
\midrule
\midrule
{\footnotesize Per-token Q.\textsubscript{$\gs=64$}}
& {4}
& {38.2\%}
& {10.54}
& {45.64}
& {20.91}
& {39.15}
& {\textbf{28.50}}
& {31.6\%}
& {27.31}
\\
{\footnotesize KCVT Quant}
& {4}
& {27.1\%}
& {12.51}
& {43.14}
& {20.91}
& {33.89}
& {24.32}
& {25.7\%}
& {26.06}
\\
{\footnotesize KIVI\textsubscript{$\gs=64,\nbu=64$}}
& {4}
& {38.2\%}
& {\textbf{13.41}}
& {48.37}
& {20.81}
& {40.98}
& {26.86}
& {31.6\%}
& {27.58}
\\
\midrule
{\footnotesize ${\textrm{\bf GEAR-L}}_{\rr=4}^{(\textrm{KCVT})}$ }
& {4}
& {30.4\%}
& {12.51}
& {47.23}
& {21.18}
& {\textbf{41.39}}
& {26.93}
& {27.3\%}
& {27.65}
\\
{\footnotesize ${\textrm{\bf\ouralg}}_{\rs=2\%,\rr=4}^{(\textrm{KCVT})}$ }
& {4}
& {32.4\%}
& {13.19}
& {\textbf{49.43}}
& {\textbf{21.28}}
& {41.32}
& {26.97}
& {29.3\%}
& {\textbf{27.80}}
\\
\midrule
\midrule
{\footnotesize Per-token Q.\textsubscript{$\gs=64$}}
& {2}
& {25.7\%}
& {0.08}
& {0.83}
& {19.78}
& {40.31}
& {25.50}
& {17.5\%}
& {27.69}
\\
{\footnotesize KIVI\textsubscript{$\gs=32,\nbu=128$}}
& {2}
& {34.9\%}
& {12.74$^\dagger$}
& {42.54}
& {20.50$^\dagger$}
& {42.71$^\dagger$}
& {25.72}
& {19.7\%}
& {27.83}
\\
\midrule 
{\footnotesize ${\textrm{\bf GEAR-L}}_{\rr=4}^{(\textrm{KIVI},\gs=64)}$}
& {2}
& {27.5\%}
& {12.63}
& {47.01}
& {\textbf{20.69}}
& {42.35}
& {26.67}
& {19.1\%}
& {\textbf{27.90}}
\\
{\footnotesize ${\textrm{\bf\ouralg}}_{\rs=2\%,\rr=4}^{(\textrm{KIVI},\gs=64)}$}
& {2}
& {31.5\%}
& {\textbf{13.04}}
& {\textbf{49.96}}
& {20.59}
& {\textbf{43.22}}
& {\textbf{27.73}}
& {23.1\%}
& {25.48}
\\
\bottomrule
\end{tabular}
\end{small}
\end{center}
\vspace{-8mm}
\end{table*}
}

\subsection{Main Results}\label{sec:main_experimental_results}
{\bf Generative performance on hard CoT reasoning tasks}. 
We compare different methods with LLaMA3-8B, LLaMA2-13B, and Mistral-7B on three challenging CoT generative tasks: GSM8k, AQuA, and BBH with 8-shot CoT demonstrations. GSM8k \citep{cobbe2021training} and AQuA \citep{ling2017program} are widely used math reasoning datasets that test models' ability of arithmetic reasoning. 
BBH \citep{suzgun2022challenging} is a suite of language and symbolic reasoning problems consisting of 6.5k problems within 23 subsets. 
Given the complexity of these tasks, we use the chain-of-thought prompts created by \citep{fu2023chainofthought} to improve reasoning, which contains 8-shot demonstrations of multi-step reasoning. 
With the CoT demonstrations, we have the average prefill length of GSM8k, AQuA, and BBH as  900, 1304, {1021} respectively (see Appendix~\ref{app:dataset_statistics}). We then prompt model to generate 256 tokens and extract answers from them. Therefore, our experiments involve long-sequence generation. 
Notably, as mentioned in Section~\ref{sec:introduction}, CoT prompts often contains densely correlated information across multiple reasoning steps and models need to pay close attention across steps to derive answers correctly. Hence, a relatively small compression error can be magnified along generation steps, resulting in significant deviation in model generations.


Table~\ref{tab:main_cot_result} presents experimental results on these hard CoT reasoning tasks. 
We see that {\ouralg} and GEAR-L achieves better or on par performance compared with baseline methods on all datasets and all models in both 4-bit and 2-bit compression. 
For example, in the case of 2-bit compression, {\ouralg} achieves 47.83\% average accuracy on LLaMA3-8B across three datasets, which is near-lossless compared to FP16 baseline (48.69\%) and significantly outperforms the best-performing baseline (28.82\%, KIVI). Notably, GEAR-L also establish remarkable performance -- near-lossless 4-bit compression and superior 2-bit performance compared to baselines, while demonstrating lower KV size and higher inference efficiency. Meanwhile,  as shown in Table~\ref{tab:main_cot_result} and Figrue~\ref{fig:gear_augment_quant}, regardless quantization backbone we choose, our method can always improve upon them by integrating the error-reduction techniques, showcasing its generalization ability as an efficient error-reduction framework. Thus, we highlight that {\bf GEAR(-L) is orthogonal to any off-the-shelf quantization scheme and can augment them in a plug-and-play manner to achieve near-lossless accuracy at minimal memory overheads}.

{\bf Generative performance on relatively easy tasks}. We also compare different methods on relatively easy tasks without CoT reasoning. Specifically, we evaluate the performance with LLaMA2-7B on LongBench \citep{bai2023longbench}, which is a suit of 21 long-context understanding tasks including question answering, summarization, code completion, etc. (please see Appendix~\ref{app:dataset_statistics} for task metrics and dataset statistics). The average input length of LongBench is 3642. We follow the evaluation method in \citep{bai2023longbench}, apply their evaluation metrics and report the average score across all 21 tasks. Besides, we also follow \citep{liu2024kivi} and compare the performance using LLaMA2-7B and LLaMA3-8B on GSM8k with standard 5-shot prompts. Such 5-shot demonstrations consists of 5 sampled questions and their one-step (or two-step) answers and do not involve complex CoT. Models are prompted to answer the question without multi-step reasoning, which is simpler than the setting of 8-shot CoT prompting. 

Table~\ref{tab:main_result2} present the experimental results on these relatively simpler tasks. We see that quantization methods can already achieve near-lossless 4-bit/2-bit compression on these tasks, showcasing their effectiveness on simpler tasks. For example, for 2-bit compression, per-token group-wise quantization and KIVI both yield around 27.7\% average scores across 21 tasks of LongBench. Moreover, KIVI establish near-lossless 2-bit performance on GSM8k with 5-shot standard examples for both LLaMA2-7B and LLaMA3-8B models.  After incorporating error-reduction techniques, {\ouralg} and GEAR-L can achieve better or on par performance compared to baseline quantization methods. For example, {\ouralg} achieves 49.96\% accuracy on GSM8k (5-shot) when compressing KV caches of LLaMA3-8B to 2-bit, which is 7.42\% higher then KIVI.

\begin{figure*}[t!]
	\centering
	\begin{subfigure}{0.32\textwidth}
		\centering
		\includegraphics[width=0.86\textwidth]{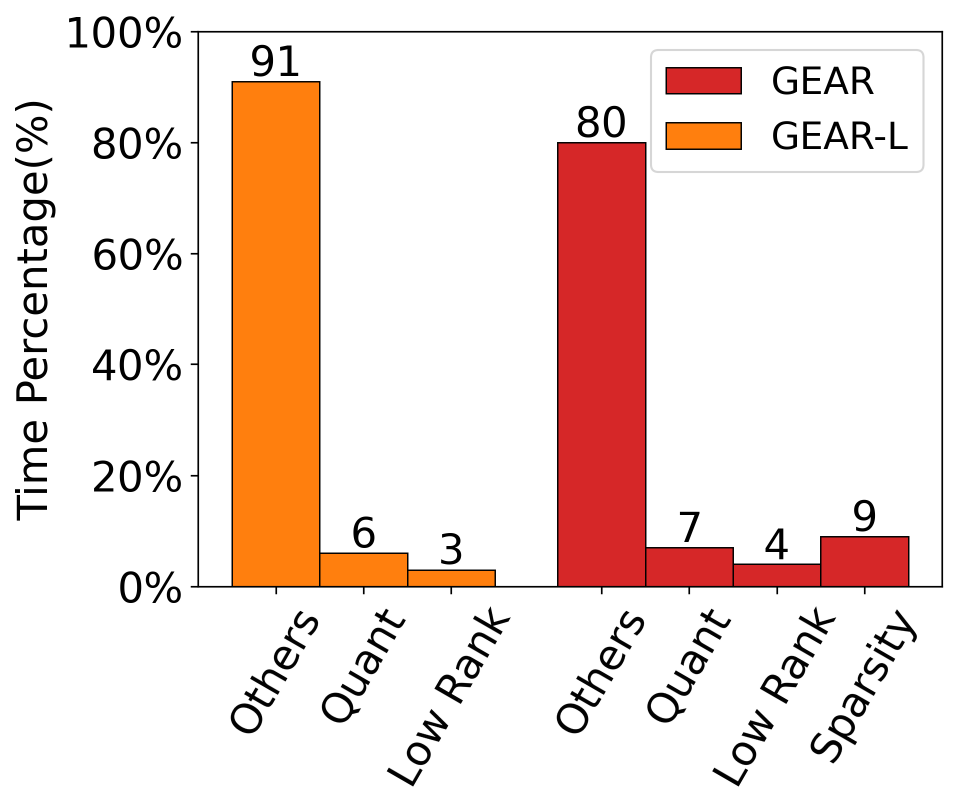}
		\caption{Time breakdown analysis}
		\label{fig:time_breakdown}
	\end{subfigure}
    \begin{subfigure}{0.32\textwidth}
		\centering
		\includegraphics[width=0.9\textwidth]{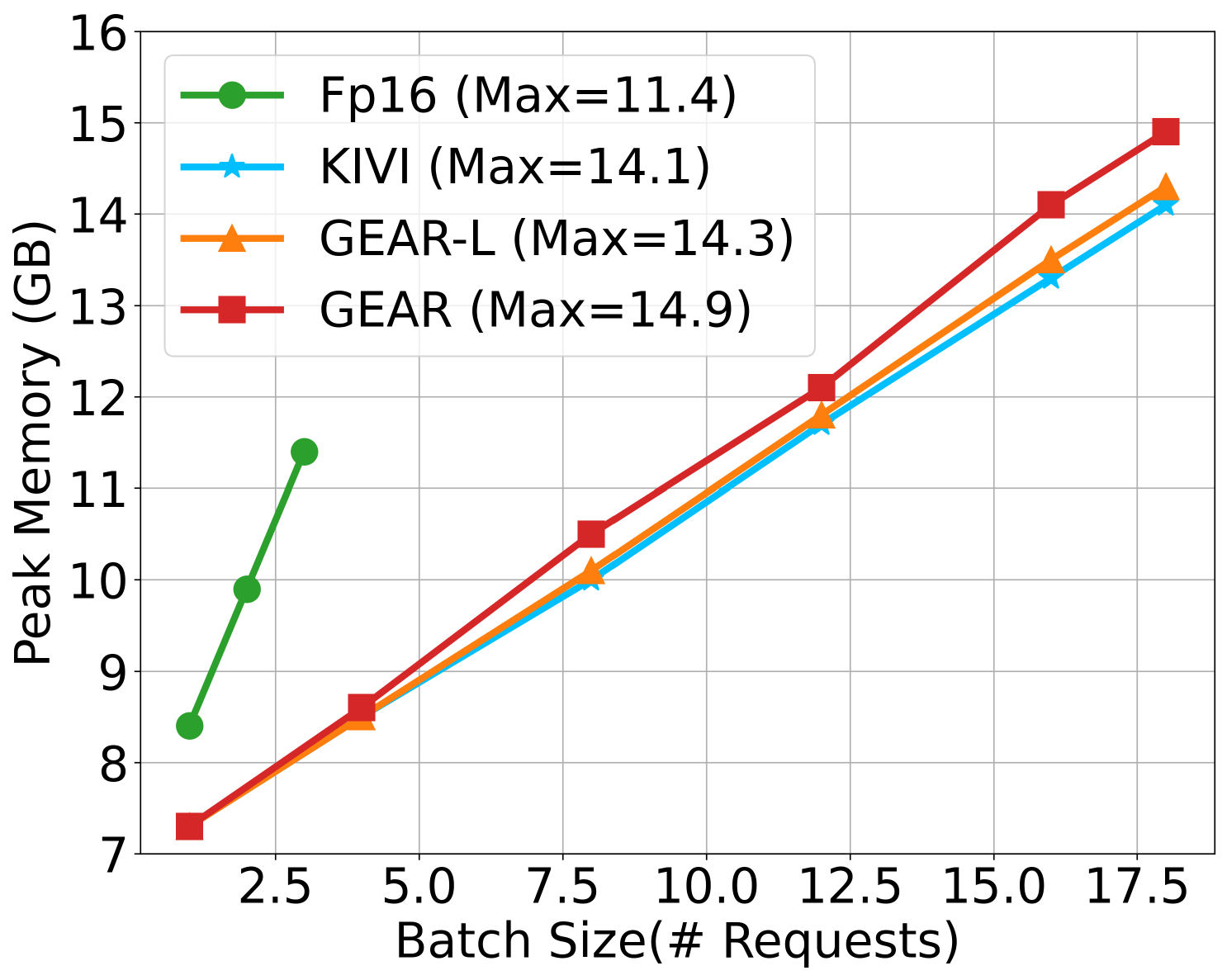}
		\caption{Peak memory comparison}
		\label{fig:peak_memory}
	\end{subfigure}
	\begin{subfigure}{0.32\textwidth}
		\centering
		\includegraphics[width=0.9\textwidth]{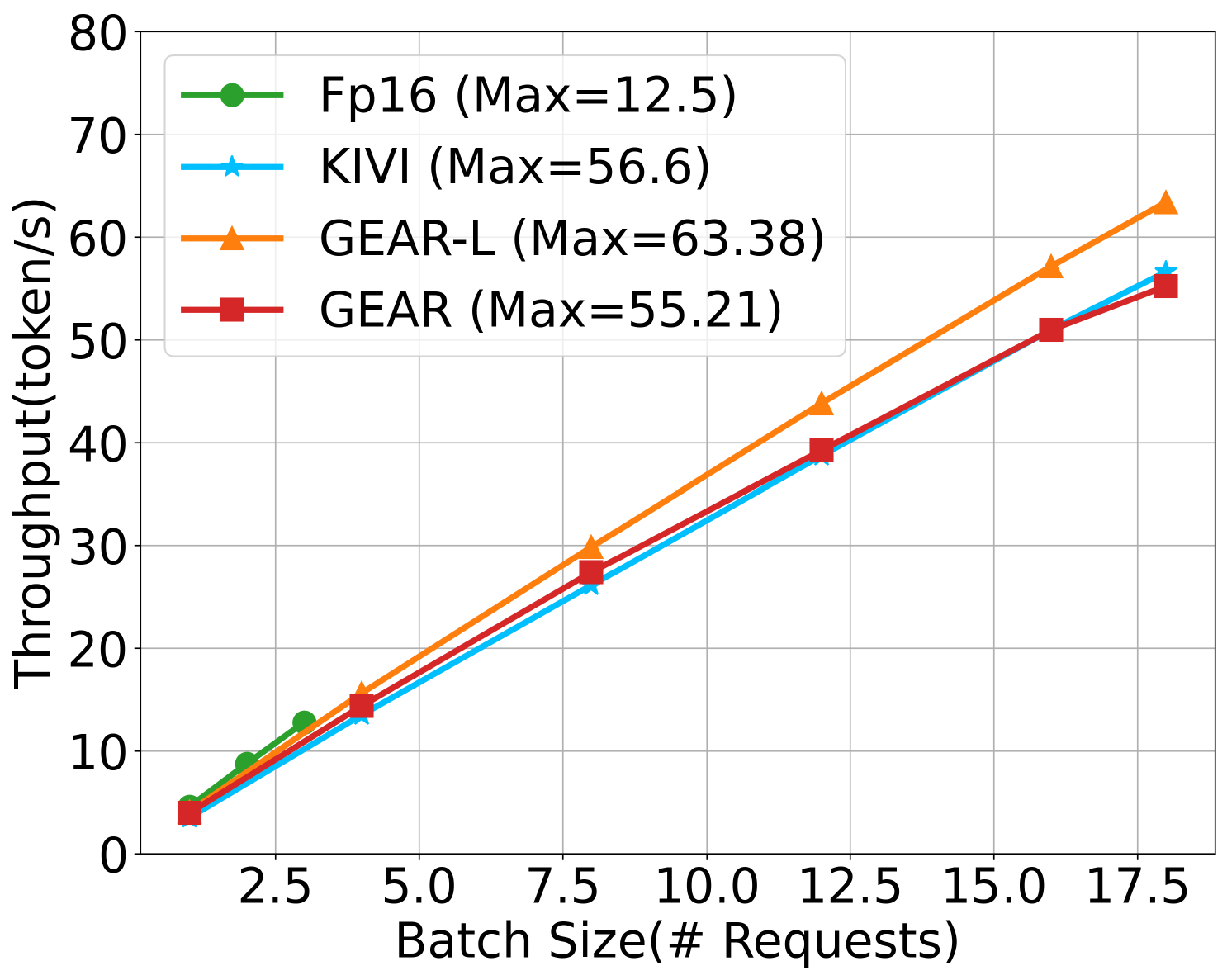}
		\caption{Throughput comparison}
		\label{fig:throughput}
	\end{subfigure}
	\caption{\small (\ref{fig:time_breakdown}) wall-clock time percentage of each component in GEAR: sparse and low-rank components induce negligible overheads. (\ref{fig:peak_memory}): GEAR significantly reduces the peak memory, enabling much larger batch size than FP16. (\ref{fig:throughput}): GEAR improve throughput significantly over FP16 due to our introduced techniques.}
	\label{fig:inference_efficiency}
\end{figure*}

\subsection{Inference Efficiency Comparison}\label{sec:inference_efficiency}
In this section, we evaluate wall-clock time, memory, and throughput of {\ouralg} on a single NVIDIA V100 GPU (16GB). Specifically, we set the input and generation length as 1000 and 500 respectively, and evaluate with LLaMA2-7B. We increase the batch size until out of memory and report the peak memory/throughput between FP16 KV caches and 2-bit quantization: KIVI, {\ouralg}, and {\ouralg}-L. We use the same hyperparameters as in Section~\ref{sec:main_experimental_results}. Here, to maximize the batch size for all methods, we compress model weights to 8-bit, using bitsandbytes from {\it Huggingface Transformers} \citep{wolf2019huggingface}. 

In this inference setting, we first provide a time breakdown analysis for {\ouralg} that compares total computational time of different components during generative inference: (i) {\it quantization-related time} that consists of total quantization and dequantization time after equipping our CUDA kernel; (ii) {\it low-rank time} that includes total time of SVD approximation by Algorithm~\ref{alg:lowrank_error} and forward pass of low-rank matrices; (iii) {\it sparsity time} that contains total computational time of outlier extraction and matrix multiplication involving $\mS$ during forward pass; (iv) {\it other time} that is primarily about model forward pass and obtained by subtracting total wall-clock time with time summation of aforementioned three items. We use the maximum batch size here (which is 18) and report the average over three trials. 
Figure~\ref{fig:time_breakdown} presents the time percentage of each component in GEAR and GEAR-L during generative inference. The results suggest that, while yielding significant performance gains, low-rank and sparse components are lightweight and do not induce unacceptable overheads. The primary complexity still stems from model forward pass. The additional latency by low-rank and sparsity components can be negligible due to our optimized implementation and inference techniques. 

Figure~\ref{fig:peak_memory} present the peak memory comparison across different batch sizes under the same inference setting. We see that, given the same batch size, GEAR significantly reduces the peak memory compared to FP16 baseline, increasing the maximum severing number (i.e.,~batch size) from 3 to 18. 
Moreover, Figure~\ref{fig:throughput} shows the throughput comparison across various batch sizes. The results demonstrate that, compared to FP16 baseline, our method significantly improves the throughput by up to $5.07\times$. Meanwhile, GEAR-L achieves slightly better throughput than KIVI due to our improved streaming strategy. We persent the detailed results of Figure~\ref{fig:peak_memory} and~\ref{fig:throughput} in Appendix~\ref{app:app_inference_efficiency}.

\subsection{Analysis and Ablation Study}\label{sec:analysis_ablation}
\begin{figure*}[t!]
	\vspace{-9mm}
	\centering
    \begin{subfigure}{0.30\textwidth}
		\centering
		\includegraphics[width=0.9\textwidth]{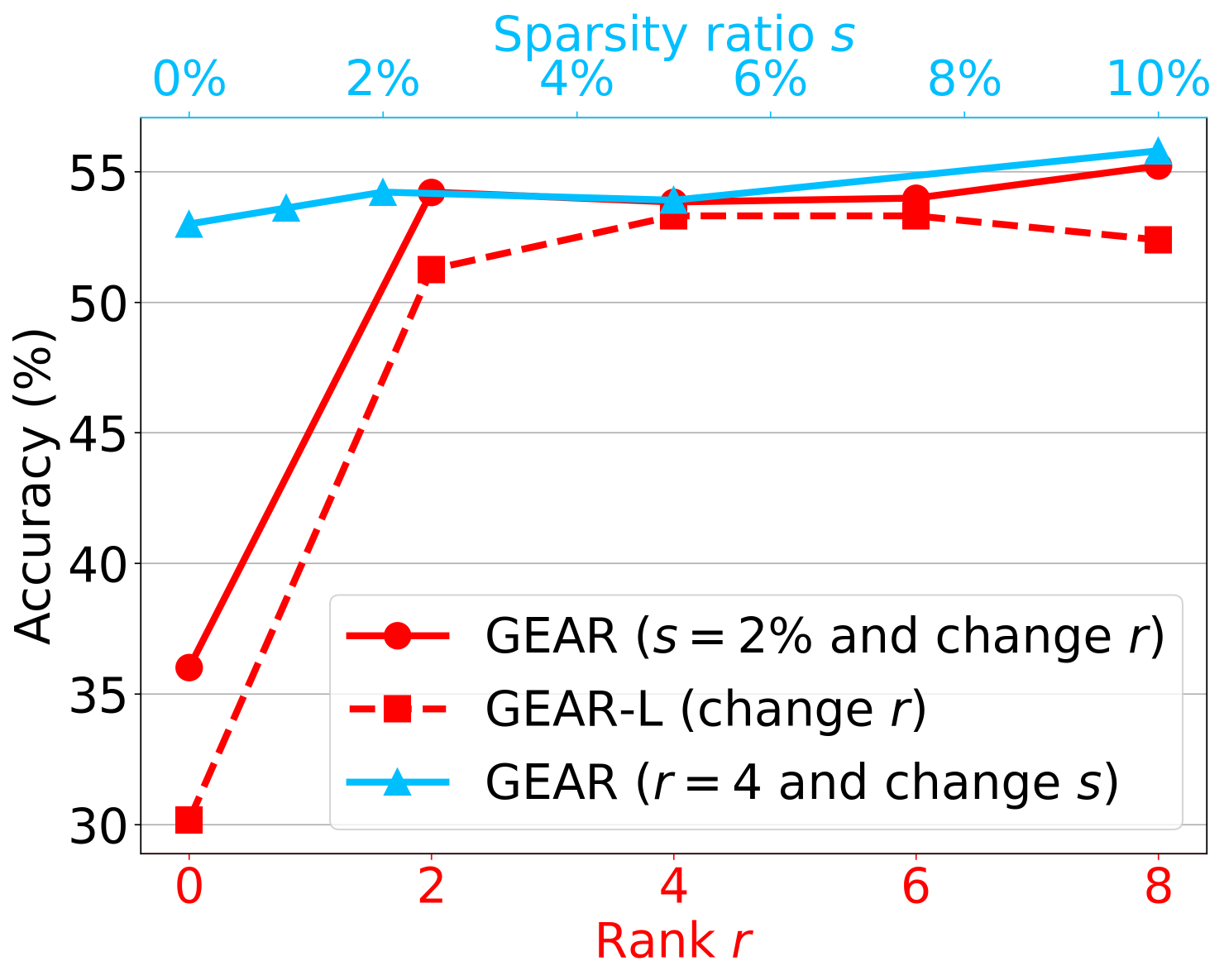}
		\vspace{-2mm}
		\caption{Ablation study on $\rs$ and $\rr$}
		\label{fig:parameter_ablation}
	\end{subfigure}
    \begin{subfigure}{0.305\textwidth}
		\centering
		\includegraphics[width=0.82\textwidth]{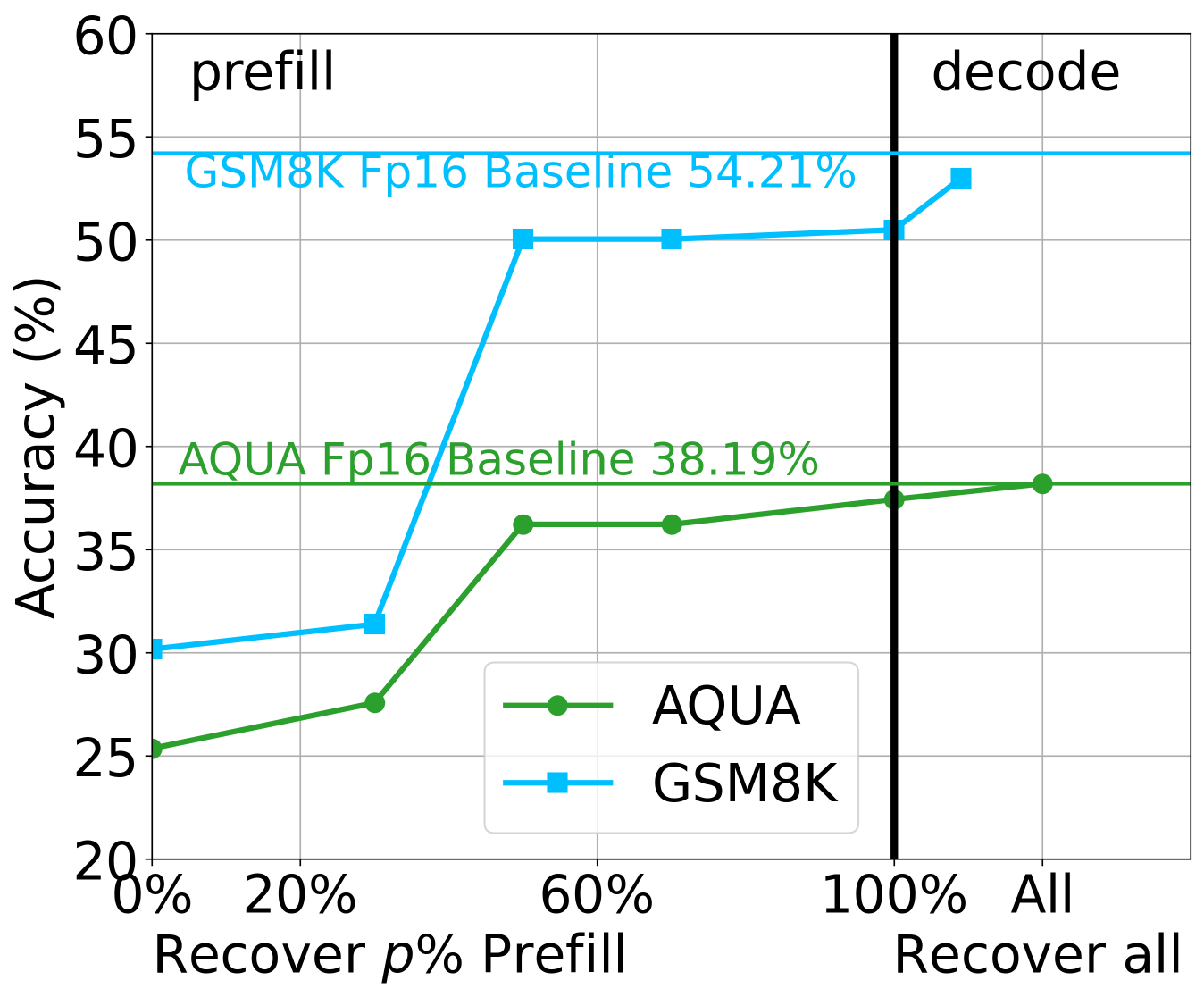}
		\vspace{-2mm}
		\caption{Recover error for $p\%$ tokens}
		\label{fig:recover_error_ablation}
	\end{subfigure}
    \begin{subfigure}{0.29\textwidth}
		\centering
		\includegraphics[width=0.90\textwidth]{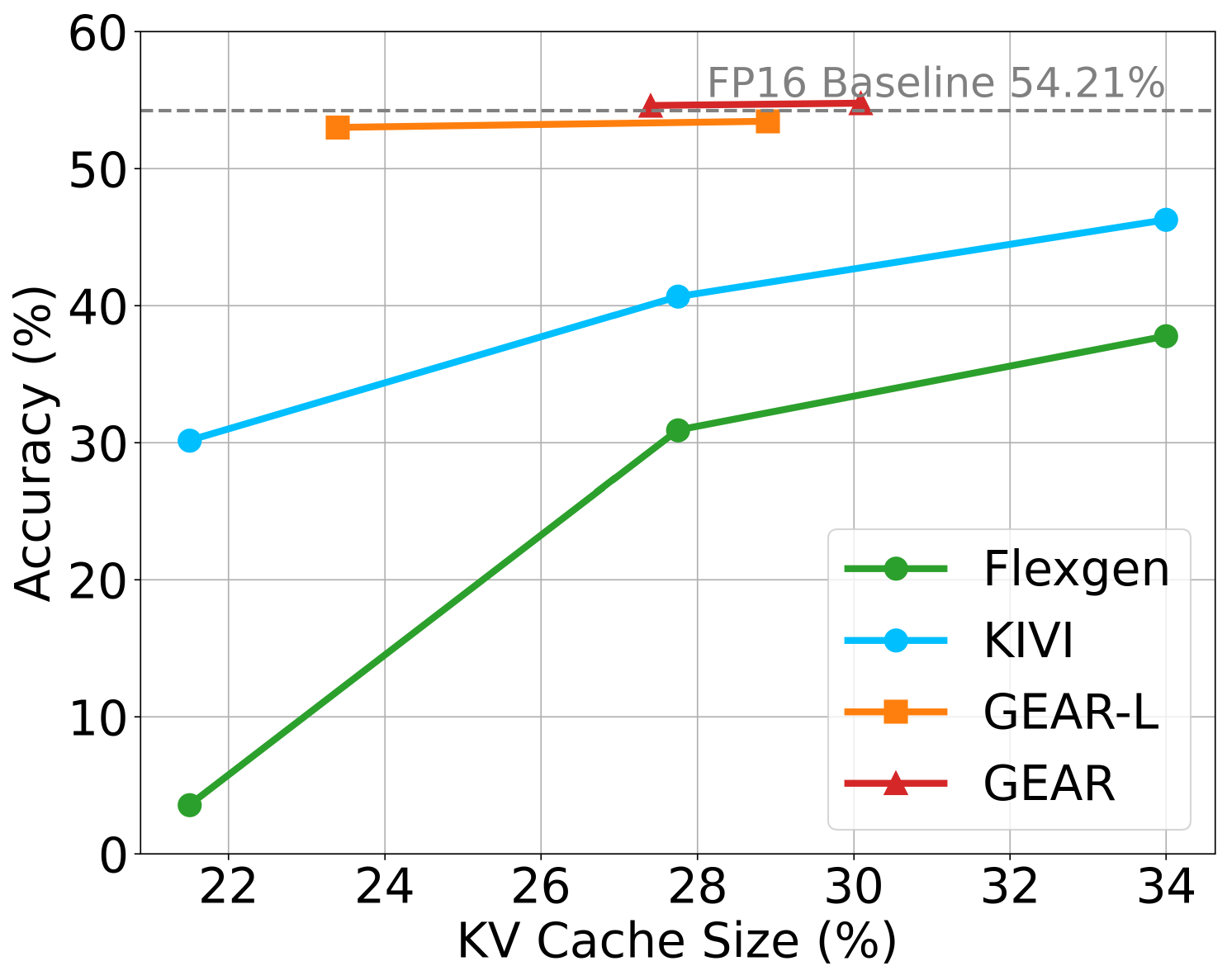}
		\vspace{-2mm}
		\caption{Acc. v.s.~KV cache size}
		\label{fig:acc_kv_size}
	\end{subfigure}
	\vspace{-2mm}
	\caption{Analysis and ablation study with LLaMA3-8B on GSM8k-CoT under 2-bit compression.}
	\label{fig:analysis_ablation}
	\vspace{-8mm}
\end{figure*}

{\bf Ablation study on sparsity ratio $\rs$ and rank $\rr$.} We study the sensitivity of GEAR to the sparsity ratio $\rs$ and rank $\rr$. Figure~\ref{fig:parameter_ablation} shows 2-bit quantization of GEAR and GEAR-L using LLaMA3-8B on GSM8k (w.~CoT) when varying $\rs$ or $\rr$. 
We see that GEAR does not require abundant sparse either low-rank components -- a small sparse ratio ($\rs=2\%$ for GEAR) and a small rank ($\rr=4$ for GEAR and GEAR-L) is adequate to achieve near-lossless 2-bit compression, demonstrating high efficiency of our method. Further increasing $\rs$ or $\rr$ may improve the accuracy but not significantly, which however results in additional memory overheads. More importantly, discarding low-rank component can significantly degenerate the performance of GEAR and GEAR-L, highlighting its vital role in error reduction. On the other hand, discarding sparse matrices can hurt performance but not significantly because the incoherent error from outlier entries can also be partially remedied by entry grouping of quantization. 
Thus, we highlight GEAR-L for those prioritizing efficiency.

{\bf The role of low-rank approximation.} We compare GEAR with outlier-aware quantization to highlight the importance of low-rank approximation. Specifically, we apply the same evaluation settings as Section~\ref{sec:main_experimental_results}. Table~\ref{tab:compare_gear_outlier} in Appendix~\ref{app:compare_gear_outlier} presents the 2-bit performance of outlier-aware KIVI quantization. The results suggest that employing outlier extraction alone for quantization can improve the performance but cannot achieve near-lossless 2-bit performance that GEAR does. 
Outlier-aware quantization still faces challenges in achieving high-ratio compression. In contrast, low-rank approximation plays a pivotal role in fully remedy approximation errors and achieving near-lossless high-ratio compression. 

{\bf Applying error reduction to different amounts of tokens.} To further demonstrate the effectiveness of error reduction, we study the performance variation of GEAR-L when applying low-rank approximation to varying number of tokens with LLaMA3-8B on GSM8k and AQuA (w.~CoT). 
Specifically, we split tokens into (i) input tokens in prefill phrase and (ii) generated tokens in decoding phrase. By default, we recover quantization errors for all of them. Alternatively, we can take $p\%$ most recent prefill tokens and only apply low-rank approximation to them. 
Figure~\ref{fig:recover_error_ablation} presents the performance of GEAR-L when changing $p$. We see that the performance of GEAR-L degenerates when decreasing the number of token applied error reduction, validating the effectiveness of error-reduction technique.

{\bf Different compression ratios.} Figure~\ref{fig:acc_kv_size} compares the performance of various methods on GSM8k (w.~CoT) when compressing KV caches of LLaMA3-8B to different remaining size. We see that GEAR and GEAR-L consistently outperform other quantization baseline methods, achieving near-lossless accuracy across various compression ratios and showcasing their effectiveness as an efficient error-reduction framework for KV cache quantization.

\section{Related Work}\label{sec:related_work}

{\bf LLM weights compression.}
LLM weight compression can significantly reduce the memory footprint and data transfer cost. GPTQ \cite{frantar2023gptq} accelerated the optimal brain quantization for LLM weights by orders of magnitude.
SqueezeLLM \cite{kim2023squeezellm} successfully compressed the model weights to 3 bits by extracting the outlier values and quantize the remaining values according to hessian matrix within 10\% perplexity increases.  These algorithms are effective and could compress weights to 2  or 3 bits with acceptable loss of accuracy. However, these methods often require significant latency overhead and gradient information to work. Thus their are not fit for KV cache compression since KV cache does not have any trainable parameter and changes every generation stage, requiring efficient light-weight method for online compression.

{\bf LLM KV cache compression.}
Activation and KV cache compression are harder than weight compression since they are more sensitive and related to model inputs. SmoothQuant \cite{xiao2023smoothquant} achieved 8-bit compression both for activation (KV caches included) and weights by adjusting the scaling factors to reduce outlier error and demonstrates near lossless performance on simple generative tasks. 
Atom \cite{zhao2023atom} successfully compressed KV Cache to 4 bits on simple generative tasks within 5\% performance degradation by combining 4-bit and 8-bit channel-wise quantization.
Another line of work explored KV pruning via token dropping based on attention score analysis. In specific, H\textsubscript{2}O \citep{zhang2023h2o} and FastGen \cite{ge2023model} proposed to prune KV via dropping tokens based on attention score to decrease the KV cache size. SparQ \cite{ribar2023sparq} not only dropped tokens according to attention score sparsity but also incorporated the error of the pruned value cache.
These pruning and quantization algorithms often work well on summarizing tasks and zero-shot inference. However, for fine-tuned models, CoT inference, and generative reasoning datasets, attention scores are denser and each token contains important information that can not be ignored. Moreover, token dropping needs to weigh each token based on attention score, which makes these methods hard to deploy with FlashAttention \cite{dao2022flashattention}. Additionally, recent works have demonstrated the attention sparsity to be a function of the non-linearity choice of the model \cite{mirzadeh2023relu}, showing its vulnerability as a metric for KV compression. 


    
\section{Discussion and Conclusions}\label{sec:discussion_conclusion}

In this paper, we present GEAR, an efficient error-reduction framework that can augment any off-the-shelf KV cache quantization scheme with two lightweight error reduction techniques in a plug-and-play manner to achieve near-lossless accuracy at high-ratio compression. In specific, GEAR demonstrates the SOTA performance on complex generative tasks involving reasoning, achieving an average accuracy improvement of $14.95\%$ at 2-bit KV quantization compared to the alternatives. Additionally, GEAR can substantially reduce the peak memory compared to FP16 baseline, and thus enable to serve more inference requests, bringing up to $\sim 5.07\times$ throughput improvement. 

\subsection{Limitations and Broader Societal Impact}
A potential limitation of our work is that we set the rank identical for every Key/Value matrix when recover quantization residuals with low-rank approximation. This ignores the fact that the importance of Key/Value matrices varies significantly across layers and heads \cite{zhang2023adaptive}. Empirically, we find that it can further improve the performance of GEAR by adaptively allocating the budget of low-rank approximation across various Key and Value matrices. We leave it as a future exploration. 

We consider our work to have direct benefit in reducing inference energy and thus carbon footprint for LLM serving, enabling democratization of the power of generative AI. At the same time, our work takes one step towards making LLM deployment efficient across various hardware platforms, increasing the user responsibility to explore the true power of AI in serving human life. We hope our method can open a new avenue of memory-efficient LLM inference for complex generation serving.




\bibliography{main}
\bibliographystyle{ims}

\section{Detailed Algorithm of {\ouralg}}\label{app:gear_algorithm}
{
\begin{algorithm}[htb!]
	\caption{{\ouralg}} 
	\label{alg:our_algorithm}
	\begin{algorithmic}[1]
		\STATE {{\bf Input:} The initial $ \{\mK_{0}, \mV_{0}\} $ of each layer, the sparsity ratio $ \rs $, the bit-width $ \bb $, the rank for prefill token $ \rr_p $, the rank for generated token $\rr_g$, the buffer $\Bcal$.}
        \STATE {\bf (Prefill Phase):}
        \FOR{$ \mX \in \{\mK_{0}, \mV_{0}\} $}
		\STATE Compute $ \mS = \Filter_{\rs}(\mX) $;
		\STATE Compute $ \mDq = \Quant_{\bb}(\mX-\mS) $;
        \STATE Compute $\mR = \mX - \mDq - \mS $;
        \FOR{$h = 1,\dots, \nh$}
		\STATE Compute $ \mLh = \SVDSolver_{\rr_p}(\mRh) $;
        \ENDFOR
        \STATE Concatenate $\mL = \textrm{Concat}(\mL_1, \dots, \mL_{\nh})$;
		\STATE Replace $ \mX $ with $ \mDq + \mL + \mS $. 
		\ENDFOR
        \STATE {\bf (Decoding Phase):}
		\FOR {$ t=1,\dots, \nng $}
		\IF{$ t \mod \nbu = 0 $}
		\FOR{$ \mX \in \{\mK_{\Bcal}, \mV_{\Bcal}\} $}
		\STATE Compute $ \mS = \Filter_{\rs}(\mX) $;
		\STATE Compute $ \mDq = \Quant_{\bb}(\mX-\mS) $;
        \FOR{$h = 1,\dots, \nh$}
		\STATE Compute $ \mLh = \SVDSolver_{\rr_g}(\mX - \mDq - \mS) $;
        \ENDFOR
        \STATE Concatenate $\mL = \textrm{Concat}(\mL_1, \dots, \mL_{\nh})$;
		\STATE Replace $ \mX $ with $ \mDq + \mL + \mS $.
		\ENDFOR
        \STATE Append $ \mK_{t} = \mK_{t-\nbu} \| \mK_{\Bcal}, \mV_{t} = \mV_{t-\nbu} \| \mV_{\Bcal} $. 
		\ELSE
		\STATE Generate new token $ \bxt $ and Push $\bkt$ to $\mK_{\Bcal}$ and Push $\bvt$ to $\mV_{\Bcal}$.
		\ENDIF
		\ENDFOR
	\end{algorithmic}
\end{algorithm}
}

\section{Power Iteration Algorithm as SVDSolver}\label{app:PI_algorithm}

The power iteration algorithm is presented in Algorithm~\ref{alg:lowrank_error}. 
\begin{algorithm}[h]
\caption{Low rank approximation of the error tensor}
\label{alg:lowrank_error}
\begin{algorithmic}
\REQUIRE Input matrix $\mX\in\R^{\nn\times\dd}$
loop iteration $L$,
low rank fraction $r$.
\STATE {\bf Output:} $\mA\in\R^{\nn\times\rr}, \mB\in\R^{\dd\times\rr}, \mA\mB^{\top}=\mL$ 
\STATE 
$\texttt{random\_initialize}(\mA)$,\\ $\texttt{random\_initialize}(\mB)$
\WHILE{$l < L$}
\IF{$l == L-1$}
\STATE $\mB \leftarrow \texttt{QRdecompostion}(\mB)$
\ENDIF
\STATE $\mA = \mX \mB$
\IF{$l == L-1$}
\STATE $ \mA \leftarrow \texttt{QRdecompostion}(\mA)$
\ENDIF
\STATE $\mB = \mX^T \mA$
\STATE $l \leftarrow l + 1$
\ENDWHILE
\end{algorithmic}
\end{algorithm}

\section{More Discussion on Related Works}\label{app:related_work}

{\bf LLM weights compression.}
LLM weight compression can significantly reduce the memory footprint and data transfer cost. GPTQ \cite{frantar2023gptq} accelerated the optimal brain quantization for LLM weights by orders of magnitude.
SqueezeLLM \cite{kim2023squeezellm} successfully compressed the model weights to 3 bits by extracting the outlier values and quantize the remaining values according to hessian matrix within 10\% perplexity increases.  These algorithms are effective and could compress weights to 2  or 3 bits with acceptable loss of accuracy. However, these methods often require significant latency overhead and gradient information to work. Thus their are not fit for KV cache compression since KV cache does not have any trainable parameter and changes every generation stage, requiring efficient light-weight method for online compression.

{\bf LLM KV cache compression.}
Activation and KV cache compression are harder than weight compression since they are more sensitive and related to model inputs. SmoothQuant \cite{xiao2023smoothquant} achieved 8-bit compression both for activation (KV caches included) and weights by adjusting the scaling factors to reduce outlier error and demonstrates near lossless performance on simple generative tasks. 
Atom \cite{zhao2023atom} successfully compressed KV Cache to 4 bits on simple generative tasks within 5\% performance degradation by combining 4-bit and 8-bit channel-wise quantization.
Another line of work explored KV pruning via token dropping based on attention score analysis. In specific, H\textsubscript{2}O \citep{zhang2023h2o} and FastGen \cite{ge2023model} proposed to prune KV via dropping tokens based on attention score to decrease the KV cache size. SparQ \cite{ribar2023sparq} not only dropped tokens according to attention score sparsity but also incorporated the error of the pruned value cache.
These pruning and quantization algorithms often work well on summarizing tasks and zero-shot inference. However, for fine-tuned models, CoT inference, and generative reasoning datasets, attention scores are denser and each token contains important information that can not be ignored. Moreover, token dropping needs to weigh each token based on attention score, which makes these methods hard to deploy with FlashAttention \cite{dao2022flashattention}. Additionally, recent works have demonstrated the attention sparsity to be a function of the non-linearity choice of the model \cite{mirzadeh2023relu}, showing its vulnerability as a metric for KV compression. 


\section{Dataset Statistics}\label{app:dataset_statistics}

Here, we show the statistics of all datasets including input length in prefill phrase, generation length and the number of evaluation examples. 

\begin{table}[h!]
    \centering
    \vspace{-2mm}
    \caption{Statistics of GSM8k, AQuA and BBH.}\label{tab:app_cot_dataset}
    \begin{tabular}{l|ccc}
    \toprule
    & \# Evaluation Example & Prefill Lenght & Generation Length \\
    \midrule
    GSM8k with 8-shot CoT & 1319 & 900 & 256 \\ 
    AQuA with 8-shot CoT & 254  & 1304 &  196\\ 
    BBH with 3-shot CoT &  6511& 1021 & 196 \\ 
    GSM8k with 5-shot examples & 1319 & 672 & 96 \\
    \bottomrule
    \end{tabular}
\end{table}

\begin{table}[h!]
    \centering
    \vspace{-2mm}
    \caption{Statistics of LongBench.}\label{tab:app_longbench}
    \begin{tabular}{l|ccc}
    \toprule
    & \# Evaluation Example & Prefill Lenght & Generation Length \\
    \midrule
    LongBench (Ave.) & 4750& 3642 & 256  \\
    \bottomrule
    \end{tabular}
\end{table}

{
\setlength{\tabcolsep}{0.21em}
\renewcommand{\arraystretch}{1.1}
\begin{table}[t!]
\centering
\caption{An overview of the dataset statistics in LongBench from \cite{bai2023longbench}. }
\begin{tabular}{lrlrccc}
\toprule
Dataset & ID & Source & Avg len & Metric & Language & \#data \\
\midrule
Single-Document QA & & & & & & \\
NarrativeQA & $1-1$ & Literature, Film & 18,409 & F1 & English & 200 \\
Qasper & $1-2$ & Science & 3,619 & F1 & English & 200 \\
MultiFieldQA-en & $1-3$ & Multi-field & 4,559 & F1 & English & 150 \\
MultiFieldQA-zh & $1-4$ & Multi-field & 6,701 & F1 & Chinese & 200 \\
\midrule Multi-Document QA & & & & & & \\
HotpotQA & $2-1$ & Wikipedia & 9,151 & F1 & English & 200 \\
2WikiMultihopQA & $2-2$ & Wikipedia & 4,887 & F1 & English & 200 \\
MuSiQue & $2-3$ & Wikipedia & 11,214 & F1 & English & 200 \\
DuReader & $2-4$ & Baidu Search & 15,768 & Rouge-L & Chinese & 200 \\
\midrule Summarization & & & & & & \\
GovReport & $3-1$ & Government report & 8,734 & Rouge-L & English & 200 \\
QMSum & $3-2$ & Meeting & 10,614 & Rouge-L & English & 200 \\
MultiNews & $3-3$ & News & 2,113 & Rouge-L & English & 200 \\
VCSUM & $3-4$ & Meeting & 15,380 & Rouge-L & Chinese & 200 \\
\midrule Few-shot Learning & & & & & & \\
TREC & $4-1$ & Web question & 5,177 & Accuracy (CLS) & English & 200 \\
TriviaQA & $4-2$ & Wikipedia, Web & 8,209 & F1 & English & 200 \\
SAMSum & $4-3$ & Dialogue & 6,258 & Rouge-L & English & 200 \\
LSHT & $4-4$ & News & 22,337 & Accuracy (CLS) & Chinese & 200 \\
\midrule Synthetic Task & & & & & & \\
PassageCount & $5-1$ & Wikipedia & 11,141 & Accuracy (EM) & English & 200 \\
PassageRetrieval-en & $5-2$ & Wikipedia & 9,289 & Accuracy (EM) & English & 200 \\
PassageRetrieval-zh & $5-3$ & C4 Dataset & 6,745 & Accuracy (EM) & Chinese & 200 \\
\midrule Code Completion & & & & & & \\
LCC & $6-1$ & Github & 1,235 & Edit Sim & Python/C\#/Java & 500 \\
RepoBench-P & $6-2$ & Github repository & 4,206 & Edit Sim & Python/Java & 500 \\
\bottomrule
\end{tabular}
\vspace{-3mm}
\end{table}
}

\section{More Inference Analysis Comparison}
\label{app:app_inference_efficiency}

\subsection{Detailed results on a single V100 GPU}

Table~\ref{tab:app_v100_efficiency} shows detailed results of inference efficiency comparison in Section~\ref{sec:inference_efficiency}, which is on a single NVIDIA V100. 
Also, to measure the peak memory save-up, we measure the memory consumption under the same batch size for both GEAR and FP16 KV cache baseline, which is 18 (the maximum batch size of GEAR on V100 GPU). Then, we apply the same inference setting and batch size for FP16 KV cache baseline and test its corresponding memory consumption on a GPU with larger GPU memory that accommodate more batches. The results shows that GEAR can reduce the memory up to $2.39\times$ compared to FP16 KV cache baseline.

\begin{table}[h!]
\vspace{-5mm}
\caption{Detailed results in Section~\ref{sec:inference_efficiency} using a single NIVIDA V100 GPU.}
\label{tab:app_v100_efficiency}
\begin{tabular}{c|c|c|c|c}
\toprule
\multicolumn{1}{l|}{Method} & \multicolumn{1}{l|}{Batch Size} & \multicolumn{1}{l|}{Time (s)} & \multicolumn{1}{l|}{Peak Memory (GB)} & \multicolumn{1}{l}{Throughputs (token/s)} 
\\ 
\midrule
& 1                               & 117                       & 8.44                                  & 4.27                                      \\
FP16                        & 2                               & 118                       & 9.94                                  & 8.47                                      \\
& 3 (max)                               & 120                       & 11.44                                 & 12.5                                      
\\ 
\midrule
& 1                               & 142                       & 7.28                                  & 3.52                                      \\
& 4                               & 148                       & 8.49                                  & 13.51                                     \\
KIVI-2bit                   & 8                               & 153                       & 10.10                                 & 26.14                                     \\
& 12                              & 155                       & 11.71                                 & 38.71                                     \\
& 16                              & 157                       & 13.32                                 & 50.96                                     \\
& 18 (max)                       & 159                       & 14.11                                 & 56.6                                      
\\ 
\midrule
& 1                               & 122                       & 7.28                                  & 4.1                                       \\
& 4                               & 128                       & 8.53                                  & 15.63                                     \\
GEARL-2bit                  & 8                               & 134                       & 10.13                                 & 29.85                                     \\
& 12                              & 137                       & 11.76                                 & 43.8                                      \\
& 16                              & 140                       & 13.37                                 & 57.14                                     \\
& 18 (max)                            & 142                       & 14.16                                 & 63.38                                     
\\ 
\midrule
& 1                               & 126                       & 7.31                                  & 3.97                                      \\
& 4                               & 139                       & 8.64                                  & 14.38                                     \\
GEARL-2bit                  & 8                               & 146                       & 10.53                                 & 27.4                                      \\
& 12                              & 153                       & 12.06                                 & 39.22                                     \\
& 16                              & 157                       & 14.07                                 & 50.95                                     \\
& 18 (max)                       & 163                       & 14.63                                 & 55.21                                     
\\ 
\bottomrule
\end{tabular}
\vspace{-3mm}
\end{table}

\subsection{Inference Efficiency Comparison on a RTX Titan GPU}

\begin{figure*}[htb!]
	\vspace{-1mm}
	\centering
	\begin{subfigure}{0.28\textwidth}
		\centering
		\includegraphics[width=0.95\textwidth]{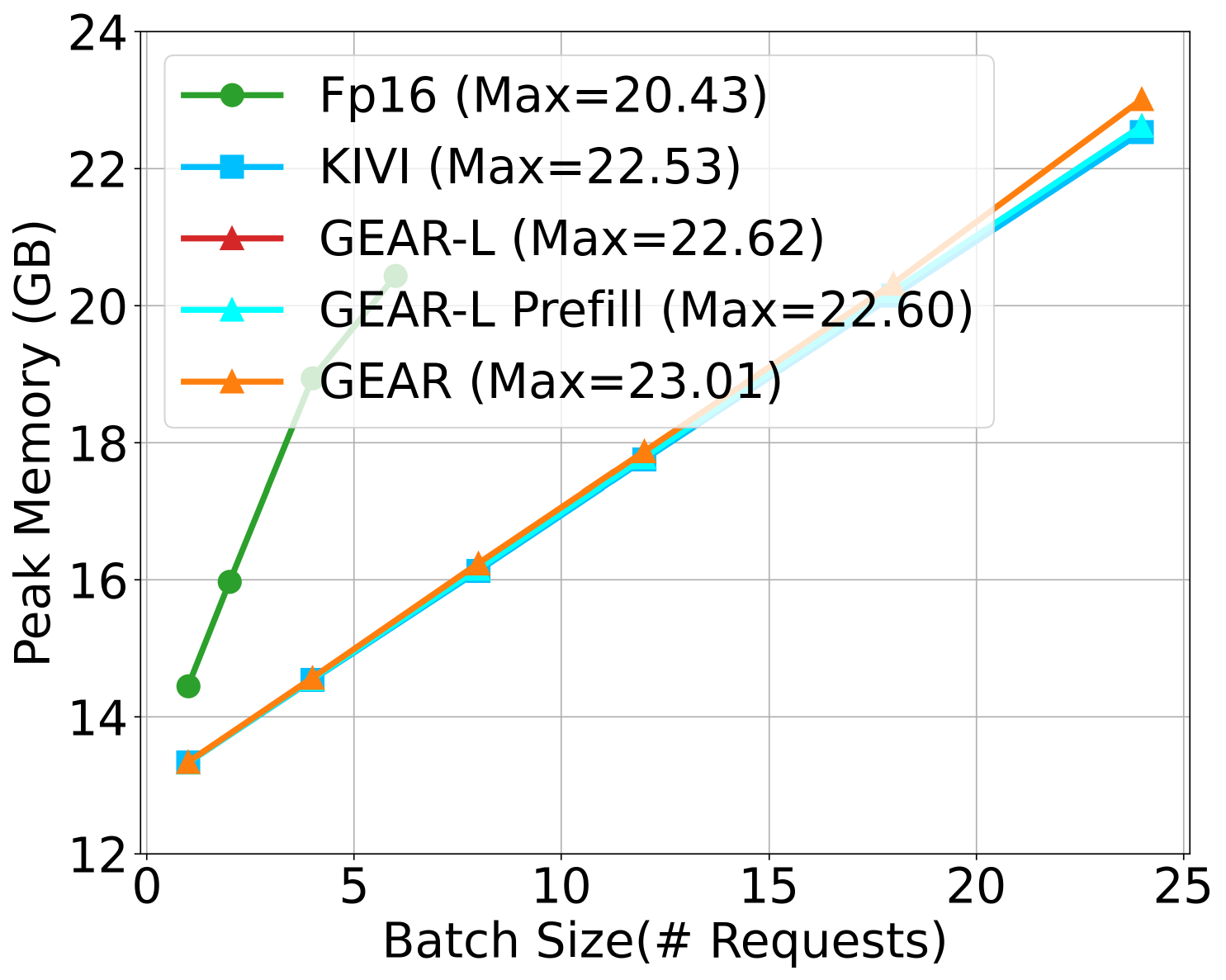}
		\vspace{-1mm}
		\caption{\small Memory Usage Comparison}
		\label{fig:RTX_Memory}
	\end{subfigure}
    \hspace{3mm}
    \begin{subfigure}{0.28\textwidth}
		\centering
		\includegraphics[width=0.95\textwidth]{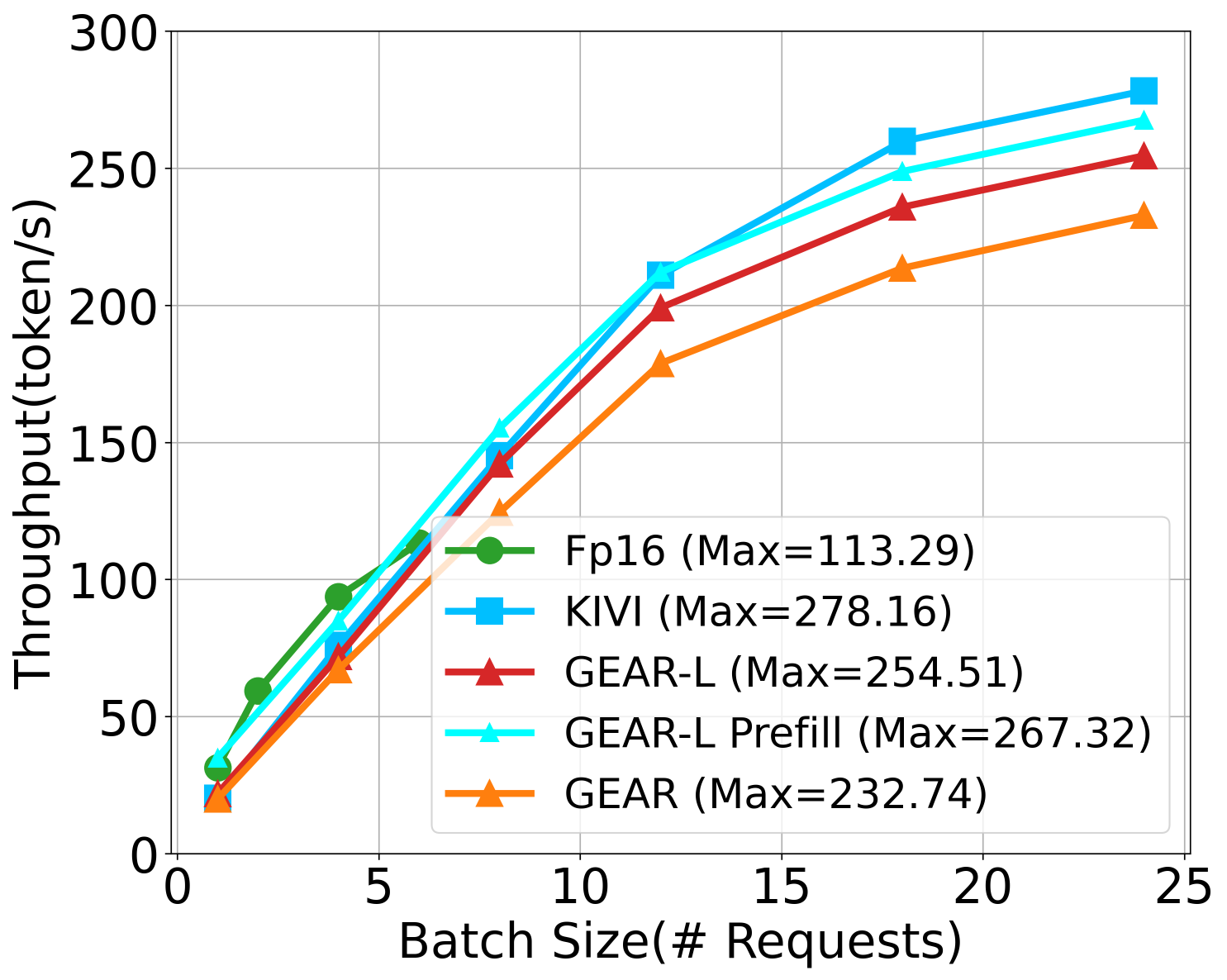}
		\vspace{-2mm}
		\caption{\small Throughput Comparison}
		\label{fig:RTX_Throughputs}
	\end{subfigure}
	\vspace{-2mm}
	\caption{\small Peak memory and throughput comparison with LLaMA2-7b on an RTX Titan 24GB GPU. }
	\label{fig:RTX}
	\vspace{-3mm}
\end{figure*}

To futher evaluate the thoughput and memory usage of GEAR, we only apply GEAR-L,GEAR-L Prefill and GEAR on a RTX Titan GPU with 24GB memory. We choose LLaMA2-7b as our base model. GEAR-L Prefill is an lite version of GEAR-L that only apply error reduction algorithm to prefill tokens. In Section~\ref{sec:analysis_ablation}, we discuss the accuracy improved by GEAR-L Prefill compared with KIVI. Here we present the Peak Memory and throughputs comparison in Figrue~\ref{fig:RTX}. With larger GPU memory, GEAR-L Prefill, GEAR-L and GEAR add acceptable latency and achieves 2.10$\times$ throughput improvement compared to Fp16 baseline. 

\subsection{KV Cache Component}

\begin{figure*}[htb!]
    \vspace{-2mm}
    \centering
    \begin{subfigure}{\textwidth}
        \centering
        \includegraphics[width=0.95\textwidth]{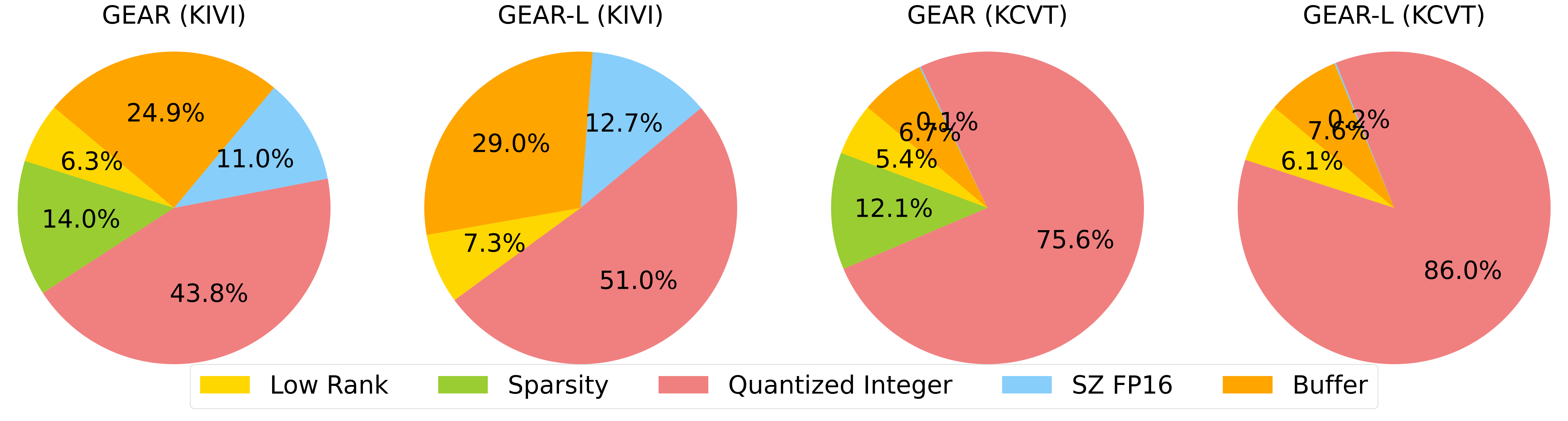}
        \vspace{-1mm}

        \label{fig:KV_Cache_KIVI}
    \end{subfigure}
    \vspace{-3mm}
    \caption{\small KV Cache memory distribution for Mistral-7B on GSM8K-CoT task}
    \label{fig:Component}
    \vspace{-3mm}
\end{figure*}

Here we discuss the components of KV Cache. Every quantization backbone at least contains quantized integer and scale\&zero point (we refer this as SZ FP16 in \autoref{fig:Component}). The size of former one is decided by quantization bit-width and the latter one is decided by group number of quantization algorithm. Another component is from the streaming buffer of FP16 residual tokens. When GEAR or GEAR-L combine with KCVT quantization, buffer size can be small. When combining with KIVI, buffer size should be larger than group size. GEAR and GEAR-L also have overheads stemming from sparsity and low rank components. 
From \autoref{fig:Component}, we can tell that, KCVT induces small streaming buffer overheads due to its large group size. In contrast, due to small group size of KIVI, it induces larger residual overheads and memory consumption from scaling factors and zero-points. 

\subsection{Comparison on Maximum Sequence Length}

In this section, we compare the maximum sequence length for different methods under the same inference setting as in Section~\ref{sec:inference_efficiency}. We use a LLaMA2-7B, set the batch size as 1, and test the maximum sequence length $\nn$ for different methods. Similarly, we compress model weights to 8-bit as in Section~\ref{sec:inference_efficiency} and apply FlashAttention to save the memory usage and allow longer sequence length. The hyperparameters are the same as Section~\ref{sec:inference_efficiency}. Table~\ref{tab:max_len} presents the maximum length for FP16 and GEAR and we can see that GEAR can increase the maximum sequence length by around 2k, making previously impossible long-sequence generation feasible. 

\begin{table}[h!]
    \centering
    \vspace{-2mm}
    \caption{Maximum sequence length comparison.}\label{tab:max_len}
    \begin{tabular}{l|cc}
    \toprule
    {\bf Method} & {\bf Bit $b$} & {\bf Max Length} \\
    \midrule
    FP16 KV Cache & 16 & 5319 \\
    $\textrm{GEAR}^{(\textrm{KIVI})}_{\rs=2\%,\rr=4}$ & 2 & 7291 \\ 
    \bottomrule
    \end{tabular}
\end{table}

\section{Comparison between GEAR and Outlier-Aware Quantization}\label{app:compare_gear_outlier}

In this section, we present the comparison between GEAR and outlier-aware quantization to further demonstrate the importance of low-rank approximation. Specifically, we apply the same evaluation settings as Section~\ref{sec:main_experimental_results}. 
Table~\ref{tab:compare_gear_outlier} present the results. 

{
\setlength{\tabcolsep}{0.20em}
\renewcommand{\arraystretch}{1.1}
\begin{table*}[htb!]
\vspace{-3mm}
\caption{Comparison of GEAR with outlier-aware quantization on CoT reasoning tasks.}
\label{tab:compare_gear_outlier}
\vspace{-3mm}
\begin{center}
\begin{small}
\begin{tabular}{l|c|c|ccc|ccc|ccc}
\toprule
\multicolumn{3}{c}{\bf Model} 
& \multicolumn{3}{|c}{\bf LLaMA3-8B} 
& \multicolumn{3}{|c|}{\bf LLaMA2-13B}
& \multicolumn{3}{|c}{\bf Mistral-7B}
\\
\midrule
\multirow{2}*{\bf Method} 
& \multirow{1}*{\bf Bit} 
& \multirow{1}*{\bf KV}
& {\bf\footnotesize GSM8k} 
& {\bf\footnotesize AQuA} 
& {\bf\footnotesize BBH}
& {\bf\footnotesize GSM8k} 
& {\bf\footnotesize AQuA} 
& {\bf\footnotesize BBH}
& {\bf\footnotesize GSM8k} 
& {\bf\footnotesize AQuA} 
& {\bf\footnotesize BBH}
\\
& {\bf $\bb$}
& {\bf size}
& {Acc} 
& {Acc}
& {Acc}
& {Acc} 
& {Acc}
& {Acc}
& {Acc} 
& {Acc}
& {Acc}
\\
\midrule
{FP16} 
& {16}
& {100\%}
& {54.21}
& {38.19}
& {53.66}
& {30.34}
& {21.65}
& {40.79}
& {42.84}
& {35.04}
& {47.92}
\\
\midrule
{\footnotesize KIVI\textsubscript{$\gs=64,\nbu=64$}}
& {2}
& {21.5\%}
& {30.17}
& {25.36}
& {30.92}
& {16.60}
& {17.72} 
& {29.43}
& {23.35}
& {22.44}
& {31.28}
\\
{\footnotesize ${\textrm{Outlier-A.}}_{\rs=2\%}^{(\textrm{KIVI})}$}
& {2}
& {24.5\%}
& {36.01}
& {36.22}
& {36.59}
& {18.19}
& {18.90}
& {33.21}
& {37.64}
& {22.44}
& {36.29}
\\
\midrule 
{\footnotesize ${\textrm{\bf GEAR-L}}_{\rr=4}^{(\textrm{KIVI})}$}
& {2}
& {23.4\%}
& {52.99} 
& {38.19} 
& {\textbf{51.44}} 
& {26.61} 
& {20.87} 
& {39.44} 
& {39.27}
& {29.92}
& {46.36}
\\
{\footnotesize ${\textrm{\bf\ouralg}}_{\rs=2\%,\rr=4}^{(\textrm{KIVI},{\gs=64})}$}
& {2}
& {27.4\%}
& {\textbf{54.59}}
& {\textbf{38.19}}
& {50.30}
& {\textbf{30.27}}
& {\textbf{23.62}}
& {\textbf{39.67}}
& {\textbf{43.14}}
& {\textbf{33.96}}
& {\textbf{48.03}}
\\
\bottomrule
\end{tabular}
\end{small}
\end{center}
\vspace{-1mm}
\end{table*}
}

\section{KV Cache Average Size for Different Datasets}

Table~\ref{tab:kv_size_main} presents the detailed KV cache size comparison across different methods, models and datasets as shown in Table~\ref{tab:main_cot_result}. 

{
\setlength{\tabcolsep}{0.28em}
\renewcommand{\arraystretch}{1.1}
\begin{table*}[htb!]
\vspace{-1mm}
\caption{Average KV Cache size for different datasets and differet models. 
}
\label{tab:kv_size_main}
\vspace{-2mm}
\begin{center}
\begin{small}
\begin{tabular}{l|c|c|ccc|ccc|ccc}
\toprule
\multicolumn{3}{c}{\bf Model} 
& \multicolumn{3}{|c}{\bf LLaMA3-8B KV Cache} 
& \multicolumn{3}{|c|}{\bf LLaMA2-13B KV Cache}
& \multicolumn{3}{|c}{\bf Mistral-7B KV Cache}
\\
\midrule
\multirow{2}*{\bf Method} 
& \multirow{1}*{\bf Bit} 
& \multirow{1}*{\bf Ave.}
& {\bf\footnotesize GSM8k} 
& {\bf\footnotesize AQuA} 
& {\bf\footnotesize BBH}
& {\bf\footnotesize GSM8k} 
& {\bf\footnotesize AQuA} 
& {\bf\footnotesize BBH}
& {\bf\footnotesize GSM8k} 
& {\bf\footnotesize AQuA} 
& {\bf\footnotesize BBH}
\\
& {\bf $\bb$}
& { KV}
& {} 
& {}
& {}
& {} 
& {}
& {}
& {} 
& {}
& {}
\\
\midrule
{FP16} 
& {16}
& {100\%}
& {100\%}
& {100\%}
& {100\%}
& {100\%}
& {100\%}
& {100\%}
& {100\%}
& {100\%}
& {100\%}
\\
\midrule
{\footnotesize Per-token Q.\textsubscript{$\gs=64$}}
& {4} 
& {34.2\%}
& {35.2\%}
& {33.0\%}
& {34.4\%}
& {35.2\%}
& {33.0\%}
& {34.4\%}
& {35.2\%}
& {33.0\%}
& {34.4\%}
\\
{\footnotesize KCVT Quant}
& {4}
& {27.1\%}
& {26.7\%}
& {27.2\%}
& {27.2\%}
& {27.5\%}
& {26.7\%}
& {27.2\%}
& {27.5\%}
& {26.7\%}
& {27.2\%}
\\
{\footnotesize KIVI\textsubscript{$\gs=64,\nbu=64$}}
& {4} 
& {34.2\%}
& {35.2\%}
& {33.0\%}
& {34.4\%}
& {35.2\%}
& {33.0\%}
& {34.4\%}
& {35.2\%}
& {33.0\%}
& {34.4\%}
\\
\midrule
{\footnotesize ${\textrm{\bf GEAR-L}}_{\rr=4}^{(\textrm{KCVT})}$  }
& {4}
& {29.0\%}
& {29.7\%}
& {28.9\%}
& {29.4\%} 
& {29.3\%}
& {28.4\%}
& {29.0\%} 
& {29.3\%}
& {28.5\%}
& {29.0\%} 
\\
{\footnotesize ${\textrm{\bf\ouralg}}_{\rs=2\%,\rho=2\%}^{(\textrm{KCVT})}$ }
& {4}
& {31.0\%}
& {31.7\%}
& {30.9\%}
& {31.4\%} 
& {31.3\%}
& {30.4\%}
& {31.0\%} 
& {31.3\%}
& {30.5\%}
& {31.0\%} 
\\
\midrule
{\footnotesize Per-token Q.\textsubscript{$\gs=64$}}
& {2} 
& {21.7\%}
& {22.7\%}
& {20.5\%}
& {21.9\%}
& {22.7\%}
& {20.5\%}
& {21.9\%}
& {22.7\%}
& {20.5\%}
& {21.9\%}
\\
{\footnotesize KIVI\textsubscript{$\gs=64,\nbu=64$}}
& {2}
& {21.7\%}
& {22.7\%}
& {20.5\%}
& {21.9\%}
& {22.7\%}
& {20.5\%}
& {21.9\%}
& {22.7\%}
& {20.5\%}
& {21.9\%}

\\
\midrule 
{\footnotesize ${\textrm{\bf GEAR-L}}_{\rr=4}^{(\textrm{KIVI})}$}
& {2}
& {23.6\%}
& {25.0\%}
& {22.7\%}
& {24.1\%}
& {24.5\%}
& {22.2\%}
& {23.7\%}
& {24.5\%}
& {22.2\%}
& {23.7\%}
\\
{\footnotesize ${\textrm{\bf\ouralg}}_{\rs=2\%,\rr=4}^{(\textrm{KIVI})}$}
& {2}
& {27.6\%}
& {29.0\%}
& {26.7\%}
& {28.1\%}
& {28.5\%}
& {26.2\%}
& {27.7\%}
& {28.5\%}
& {26.2\%}
& {27.7\%}
\\
\bottomrule
\end{tabular}
\end{small}
\end{center}
\vspace{-2mm}
\end{table*}
}

\section{Comparison with token dropping.}
We evaluate the performance of H\textsubscript{2}O \citep{zhang2023h2o} for reducing KV cache size on GSM8k with LLaMA2-7B. Table~\ref{tab:gsm8k_h20_gear} presents its accuracy when dropping 50\% tokens, which suggests H\textsubscript{2}0 cannot effectively preserve the performance nor achieve high compression ratio. For complex tasks involving reasoning or long-sequence generation (such as GSM8k), models need to closely attend to most contextual information to generate correct answers. Token dropping methods, however, can make some information directly invisible, resulting in deviation in generation and degradation of performance.  

{
\begin{table}[htb!]
\vspace{-3mm}
\caption{Accuracy of H\textsubscript{2}O on GSM8k with LLaMA2-7B.}
\vspace{-1mm}
\label{tab:gsm8k_h20_gear}
\begin{center}
\begin{tabular}{c|c|c|c}
\toprule
{\bf Method} & {\bf Bit $\bb$} & {\bf KV size} & {\bf CoT Acc.} 
\\
\midrule
{FP16} & {16} & {100\%} & {16.33} 
\\
{H\textsubscript{2}O} & {16} & {50\%} & {6.82} 
\\
\midrule
{$\textrm{GEAR}_{\rs=2\%,\rr=4}^{(\textrm{KCVT})}$} & {4} & {32.4\%} & {16.14}
\\
\bottomrule
\end{tabular}
\end{center}
\vspace{-2mm}
\end{table}
}




\section{Discussion on the Prompts}


\begin{figure}[h!] 
    \centering
    \includegraphics[width=0.70\textwidth]{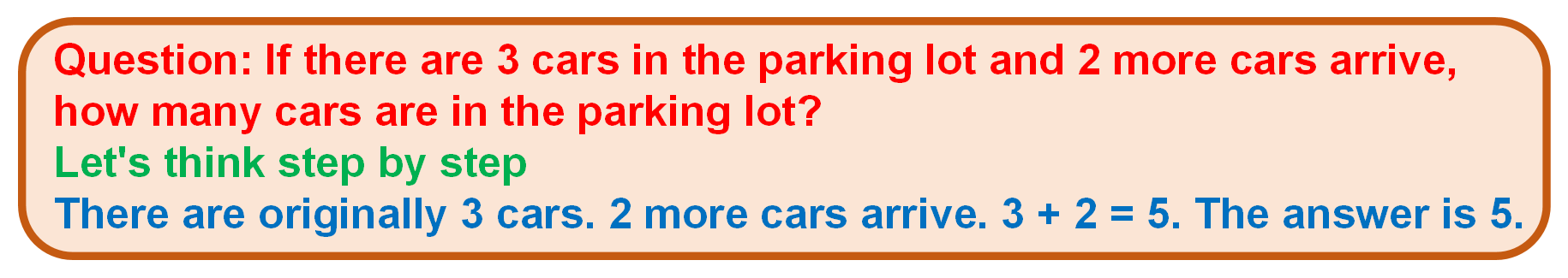}
    \vspace{0mm}
    \caption{Example of GSM8k-CoT prompt. The \textcolor{red}{Red}, \textcolor{green}{Green}, and \textcolor{blue}{Blue} colored portions correspond to the example question, a common preceding prompt, and the example answer prompt, respectively. Here, we use the common prompt to improve  the reasoning of the LLM.}
    \label{fig:cot_prompt}
    \vspace{-4mm}
\end{figure}
For the GSM8k dataset, there is a fixed prompt for all evaluations. The prompt contains 8 examples with clear guidance step by step. For the MMLU and BBH dataset, there are individual prompts for each sub dataset. \autoref{fig:cot_prompt} shows one of the example in GSM8K dataset.

\end{document}